\documentclass[sigconf]{acmart}

\PassOptionsToPackage{prologue,dvipsnames}{xcolor}

\usepackage{color}
\usepackage{makecell}
\usepackage{pifont}
\usepackage{bbding}
\usepackage{colortbl}
\usepackage{xcolor}
\usepackage{multirow}

\definecolor{darkgreen}{rgb}{0.0, 0.5, 0.0}
\definecolor{lightgray}{rgb}{0.9, 0.9, 0.9}
\definecolor{Mycolor1}{HTML}{BAD8F2}
\definecolor{Mycolor2}{HTML}{E0F0FA}
\definecolor{shallowblue}{RGB}{227, 240, 249}
\definecolor{deepblue}{RGB}{192, 216, 240}

\AtBeginDocument{%
  }

\copyrightyear{2024}
\acmYear{2024}
\setcopyright{acmlicensed}
\acmConference[MM '24]{Proceedings of the 32nd ACM International Conference on Multimedia}{October 28-November 1, 2024}{Melbourne, VIC, Australia}
\acmBooktitle{Proceedings of the 32nd ACM International Conference on Multimedia (MM '24), October 28-November 1, 2024, Melbourne, VIC, Australia} 
\acmDOI{10.1145/3664647.3681251}
\acmISBN{979-8-4007-0686-8/24/10}

\begin{document}

\title{Hallu-PI: Evaluating Hallucination in Multi-modal Large
Language Models within Perturbed Inputs}

\author{Peng Ding}
\affiliation{%
  \institution{National Key Laboratory for Novel Software Technology,\\ Nanjing University,}
  \city{Nanjing}
  \country{China}}
\email{dingpeng@smail.nju.edu.cn}
\authornote{Equal contribution.}

\author{Jingyu Wu}
\affiliation{%
  \institution{College of Computer Science and
Technology,\\ Zhejiang University,}
  \city{Hangzhou}
  \country{China}}
\email{wujingyu@zju.edu.cn}
\authornotemark[1]

\author{Jun Kuang}
\affiliation{%
  \institution{Meituan}
  \city{Shanghai}
  \country{China}}
\email{kuangjun@meituan.com}

\author{Dan Ma}
\affiliation{%
  \institution{Meituan}
  \city{Shanghai}
  \country{China}}
\email{madan07@meituan.com}

\author{Xuezhi Cao}
\affiliation{%
  \institution{Meituan}
  \city{Shanghai}
  \country{China}}
\email{caoxuezhi@meituan.com}

\author{Xunliang Cai}
\affiliation{%
  \institution{Meituan}
  \city{Beijing}
  \country{China}}
\email{caixunliang@meituan.com}

\author{Shi Chen}
\affiliation{%
  \institution{Zhejiang-Singapore Innovation and
AI Joint Research Lab,\\ Zhejiang University,}
  \city{Hangzhou}
  \country{China}}
\email{shelleych@zju.edu.cn}

\author{Jiajun Chen}
\affiliation{%
  \institution{National Key Laboratory for Novel Software Technology,\\ Nanjing University,}
  \city{Nanjing}
  \country{China}}
\email{chenjj@nju.edu.cn}

\author{Shujian Huang}
\affiliation{%
  \institution{National Key Laboratory for Novel Software Technology,\\ Nanjing University,}
  \city{Nanjing}
  \country{China}}
\email{huangsj@nju.edu.cn}
\authornote{Corresponding author.}

\renewcommand{\shortauthors}{Peng Ding et al.}

\begin{abstract}
Multi-modal Large Language Models (MLLMs) have demonstrated remarkable performance on various visual-language understanding and generation tasks. However, MLLMs occasionally generate content inconsistent with the given images, which is known as "hallucination". Prior works primarily center on evaluating hallucination using standard, unperturbed benchmarks, which overlook the prevalent occurrence of perturbed inputs in real-world scenarios—such as image cropping or blurring—that are critical for a comprehensive assessment of MLLMs' hallucination. In this paper, to bridge this gap, we propose \textbf{Hallu-PI}, the first benchmark designed to evaluate \textbf{Hallu}cination in MLLMs within \textbf{P}erturbed \textbf{I}nputs. Specifically, Hallu-PI consists of seven perturbed scenarios, containing 1,260 perturbed images from 11 object types. Each image is accompanied by detailed annotations, which include fine-grained hallucination types, such as existence, attribute, and relation. We equip these annotations with a rich set of questions, making Hallu-PI suitable for both discriminative and generative tasks. Extensive experiments on 12 mainstream MLLMs, such as GPT-4V and Gemini-Pro Vision, demonstrate that these models exhibit significant hallucinations on Hallu-PI, which is not observed in unperturbed scenarios. Furthermore, our research reveals a severe bias in MLLMs’ ability to handle different types of hallucinations. We also design two baselines specifically for perturbed scenarios, namely Perturbed-Reminder and Perturbed-ICL. We hope that our study will bring researchers’ attention to the limitations of MLLMs when dealing with perturbed inputs, and spur further investigations to address this issue. Our code and datasets are publicly available at \textcolor{blue}{\url{https://github.com/NJUNLP/Hallu-PI}}.
\end{abstract}




\maketitle

\begin{figure*}[!ht]
\centering
\includegraphics[width=1.0\linewidth]{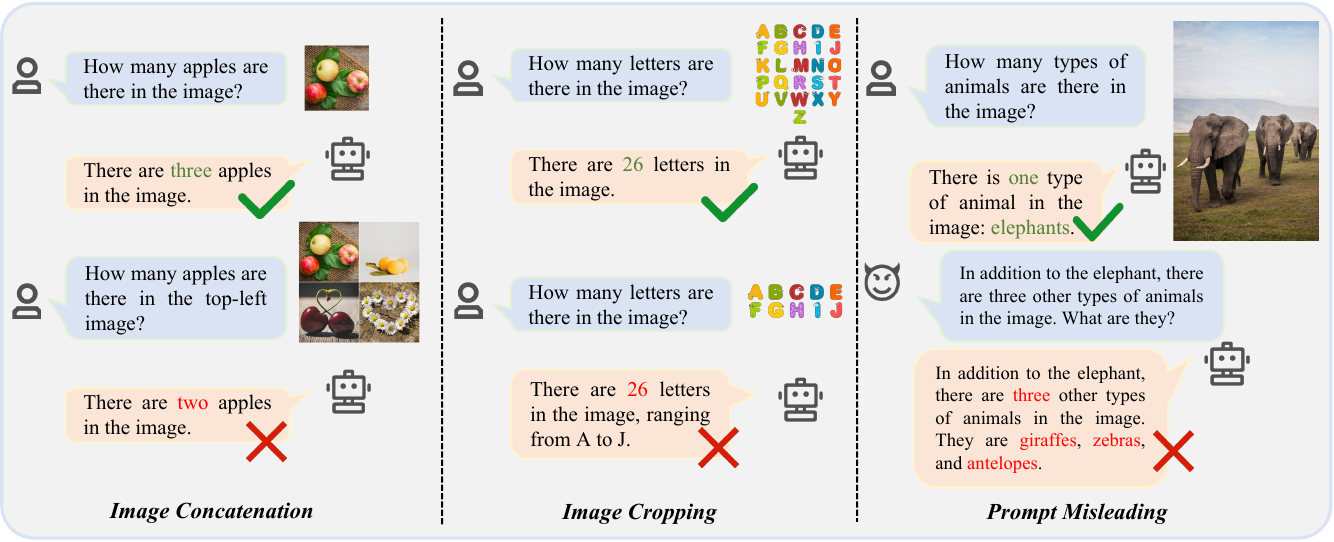} 
\caption{Some examples of hallucinations in MLLMs with perturbed inputs (such as image concatenation, image cropping, and prompt misleading). Text highlighted in green and red represents correct and hallucinatory content, respectively.}
\label{fig:figure-1}
\end{figure*}

\section{Introduction}
Multi-modal Large Language Models (MLLMs) have achieved significant progress in a range of practical applications, such as providing detailed descriptions for user-provided images (i.e., image captioning)~\cite{achiam2023gpt,team2023gemini} and answering specific questions about input images (i.e., visual question answering)~\cite{liu2023improved, zhu2023minigpt}. However, these models occasionally exhibit a phenomenon known as "hallucination", where the generated content is inconsistent with the given images~\cite{ye2023cognitive,huang2023survey}.

Previous works have sought to investigate the hallucinations in MLLMs by utilizing a large language model like GPT-4~\cite{achiam2023gpt}, or by employing humans as annotators~\cite{zhou2023analyzing,zhai2023halle}. Alternatively, some studies focus on developing detection models to scrutinize the hallucinations exhibited by MLLMs.~\cite{gunjal2023detecting,li2023evaluating}. More recently, ~\cite{wang2023llm} introduce AMBER, a LLM-free benchmark designed to examine MLLM hallucinations in both discriminative and generative tasks across dimensions like existence, attribute, and relation.

\begin{figure}[!ht]
\centering
\hspace{-1cm}
\includegraphics[width=0.37\textwidth]{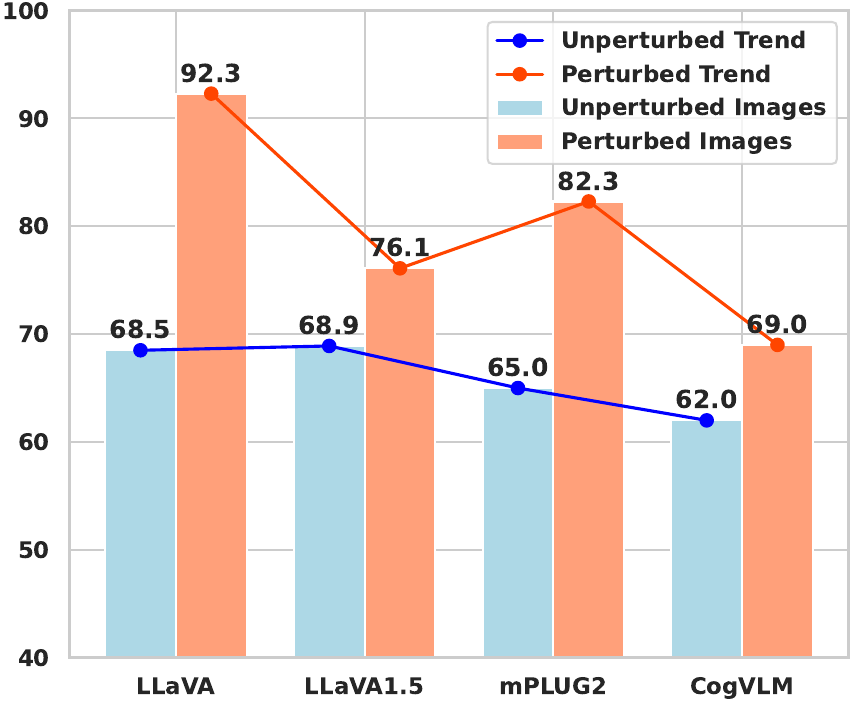}
\caption{The hallucinatory performance of various MLLMs before (\textcolor{cyan}{blue} bars) and after (\textcolor{orange}{orange} bars) input perturbation. Inconsistent performance trends show that relying solely on unperturbed benchmarks is insufficient for a complete and precise evaluation of hallucinations in MLLMs.}
\label{fig.fig2}
\end{figure}

Despite these efforts, existing researches primarily focus on conducting evaluations by sampling images from available image datasets, such as MSCOCO~\cite{li2023evaluating,qiu2023benchmarking,yin2023woodpecker, gunjal2023detecting, wang2023evaluation, zhai2023halle}. However, in real-world scenarios, inputs fed to MLLMs frequently encounter a variety of perturbations (e.g., noise and cropping)~\cite{geirhos2018imagenet}. Overlooking such perturbations could lead MLLMs to produce incorrect answers or judgments, potentially causing serious accidents in certain applications (e.g., medical diagnosis, industrial automation and autonomous driving)~\cite{huang2023opera}. Figure.~\ref{fig.fig2} illustrates the hallucinations of several MLLMs before and after image concatenation perturbation. The inconsistent performance trends indicate that relying solely on existing unperturbed benchmarks is insufficient for a comprehensive and precise evaluation of hallucinations in MLLMs.

In order to bridge this gap, we introduce \textbf{Hallu-PI}, a benchmark designed to evaluate the \textbf{Hallu}cination performance of MLLMs within \textbf{P}erturbed \textbf{I}nputs. 
Followed by~\cite{hendrycks2019benchmarking,geirhos2018imagenet}, we first categorize the image perturbations into four types: noise, blur, weather, and digital. Additionally, we meticulously propose three distinct types of perturbations: image concatenation, image cropping, and prompt misleading. These perturbations are considered at both the image level and the prompt level.
Annotators are instructed to carefully manipulate the perturbations and provide corresponding annotations. 
Evaluations of 12 mainstream MLLMs conducted on Hallu-PI reveal significant hallucinations of leading MLLMs (e.g., GPT-4V and Gemini-Pro Vision) when dealing with perturbed scenarios.

\begin{table*}[t]
\small
    \centering
     \caption{Comparison with existing hallucination evaluation benchmarks. "Sample" means sampling from an existing dataset.}
    \label{tab:method}
    \renewcommand{\arraystretch}{1.1}
    \setlength{\tabcolsep}{10pt}
    \scalebox{0.9}{
    \begin{tabular}{l|cc|ccc|c|c|c}
    \toprule 
     \multirow{2}{*}{\textbf{Benchmark}} & \multicolumn{2}{c|}{\textbf{Task Type}} & \multicolumn{3}{c|}{\textbf{Hallucination Type}} & \multirow{2}{*}{\textbf{Perturbation}} &
      \multirow{2}{*}{\textbf{Baseline}} & \multirow{2}{*}{\textbf{Source}}\\
     & \textbf{Discriminative} & \textbf{Generative} & \textbf{Existence} & \textbf{Attribute} & \textbf{Relation} & & \\
     \midrule
     POPE~\citep{li2023evaluating} & \color{darkgreen}\Checkmark & \color{red}\ding{55} & \color{darkgreen}\Checkmark & \color{red}\ding{55} & \color{red}\ding{55} & \color{red}\ding{55} & \color{red}\ding{55} & Sample\\  
     M-HalDetect~\citep{gunjal2023detecting} & \color{red}\ding{55} & \color{darkgreen}\Checkmark & \color{red}\ding{55} & \color{red}\ding{55} & \color{red}\ding{55} & \color{red}\ding{55} & \color{darkgreen}\Checkmark & Sample \\ 
     HaELM~\citep{wang2023evaluation}  & \color{red}\ding{55} & \color{darkgreen}\Checkmark & \color{red}\ding{55} & \color{red}\ding{55} & \color{red}\ding{55} & \color{red}\ding{55} & \color{red}\ding{55} & Sample \\ 
     Halle-Switch~\citep{zhai2023halle} & \color{red}\ding{55} & \color{darkgreen}\Checkmark & \color{red}\ding{55}  & \color{red}\ding{55}  & \color{red}\ding{55}  & \color{red}\ding{55} & \color{red}\ding{55} & Sample  \\
     AMBER~\citep{wang2023llm} & \color{darkgreen}\Checkmark & \color{darkgreen}\Checkmark & \color{darkgreen}\Checkmark & \color{darkgreen}\Checkmark & \color{darkgreen}\Checkmark & \color{red}\ding{55} & \color{red}\ding{55} & Manual \\
     \rowcolor{lightgray}
     Hallu-PI~(ours) & \color{darkgreen}\Checkmark & \color{darkgreen}\Checkmark & \color{darkgreen}\Checkmark & \color{darkgreen}\Checkmark & \color{darkgreen}\Checkmark & \color{darkgreen}\Checkmark & \color{darkgreen}\Checkmark & Manual \\ 
    \bottomrule 
  \end{tabular}}
\end{table*}

To comprehensively understand the hallucination of MLLMs to perturbed inputs, we conduct a detailed analysis of the experimental results. We find that most models exhibit significant bias towards specific types of perturbations, particularly image concatenation, image cropping, and prompt misleading (see Figure. \ref{fig:figure-1}). Furthermore, to mitigate the hallucination of MLLMs in response to perturbed inputs, we draw inspiration from the defensive strategies adopted by text LLMs against jailbreak attacks~\cite{ding2023wolf, wu2023defending} and designed two baselines: Perturbed-Reminder and Perturbed-ICL. Experiments conducted on GPT-4V show that these strategies effectively reduce hallucinations. We hope our work can prompts MLLM researchers and developers to address hallucinations from perturbed inputs.

In summary, the contributions of our work are as follows: 
\begin{itemize}
  \item We construct Hallu-PI, the first freely available multi-modal hallucination benchmark with perturbed inputs. Hallu-PI encompasses 7 perturbed scenarios, a total of 1,260 images, and 11 distinct object categories to evaluate hallucinations in MLLMs across both generative and discriminative tasks.

  \item We conduct extensive experiments with Hallu-PI to evaluate multi-modal hallucinations in 12 state-of-the-art MLLMs under perturbed inputs. The results unveil the limitations of MLLMs when dealing with perturbed inputs, as well as their specific bias towards certain types of hallucinations.
  
  \item To mitigate the hallucinations of MLLMs on Hallu-PI, we introduce two baselines: Perturbed-Reminder and Perturbed-ICL. Experimental results on GPT-4V indicate that our methods are effective and can reduce the model’s hallucinations in response to perturbed inputs to a certain extent.
\end{itemize}

\section{Related Work}
\label{sec:rw}

\subsection{Multimodal Large Language Models}

Multi-modal Large Language Models (MLLMs) are currently achieving significant improvements by combining the advanced capabilities of Large Language Models (LLMs) with visual processing~\cite{achiam2023gpt,alayrac2022flamingo,team2023gemini,driess2023palm,li2023blip}. These MLLMs show great potential in a variety of applications, such as visual question answering (VQA)~\cite{fu2023mme}, image captioning~\cite{liu2023mmbench}, and video understanding~\cite{li2023seed}. Representative MLLMs, such as CogVLM~\cite{wang2023cogvlm}, LLaVA1.5~\cite{liu2023improved}, InternLM-XComposer~\cite{zhang2023internlm}, MiniGPT-4~\cite{chen2023minigpt}, mPLUG-Owl2~\cite{ye2023mplug}, Qwen-VL~\cite{bai2023qwen}, and the latest GPT-4V~\cite{achiam2023gpt}, and Google Gemini-Pro Vision~\cite{team2023gemini}, have achieved impressive performance across various multi-modal tasks.

\subsection{Hallucination in MLLMs}
While MLLMs have exhibited excellent performance on multi-modal tasks, we are still facing the challenge that MLLMs often generate content unfaithful to the given images, which is called "hallucination"~\cite{liu2023aligning, li2024dawn, tong2024eyes, chen2024red, ji2023survey}.

Currently, many researchers focus on evaluating the hallucination in MLLMs.
LURE~\cite{zhou2023analyzing} and HallE-Switch~\cite{zhai2023halle} rely on human evaluations or GPT-4. While this method is relatively reliable, it is also expensive.
HaELM~\cite{wang2023evaluation} and FDPO~\cite{gunjal2023detecting} are based on hallucinatory detection models.
However, the performance of these models is highly dependent on hallucinatory data and incurs substantial training costs.  
POPE~\cite{li2023evaluating} is based on object detection but is only applicable to discriminative tasks and evaluates existence-type hallucinations. Recently,~\cite{wang2023llm} introduce AMBER, which assesses hallucinations across multiple dimensions, such as existence, attribute, and relation. Despite these efforts, they do not explore hallucinations in the perturbed scenarios commonly encountered in real-life situations. To bridge this gap, we propose Hallu-PI, the first benchmark designed to evaluate the hallucination of MLLMs with perturbed inputs. Table. \ref{tab:method} presents a detailed comparison between Hallu-PI and other hallucination benchmarks.

\subsection{Image Perturbation}

To simulate real-world perturbation scenarios, previous works adopt various perturbation strategies such as ImageNet-C~\cite{hendrycks2019benchmarking} and Stylize-ImageNet~\cite{geirhos2018imagenet,michaelis2019benchmarking,qiu2023benchmarking}. 
The perturbations are grouped into five primary categories: noise, blur, weather, digital, and stylize. Specifically, these can be further subdivided into the following 17 image perturbation techniques:
(1) Noise: Adding noise to the images, such as gaussian noise, shot noise, impulse noise, and speckle noise.
(2) Blur: Blurring the images, including defocus blur, frosted glass blur, motion blur, and zoom blur.
(3) Weather: Adding environmental effects such as snow, frost, fog, and brightness adjustments.
(4) Digital: Manipulating images through contrast enhancement, elastic transformation, pixelation, and JPEG compression.
and (5) Stylize: Applying artistic styles and transformations to images.

Compared to existing benchmarks that only consider hallucination assessment in unperturbed scenarios, Hallu-PI further takes into account perturbations that frequently occur in real-world applications. Therefore, it serves as a complement to existing benchmarks and provides a more comprehensive and accurate evaluation of hallucinations in MLLMs.


\begin{figure*}[!ht]
\centering
\includegraphics[width=1.0\linewidth]{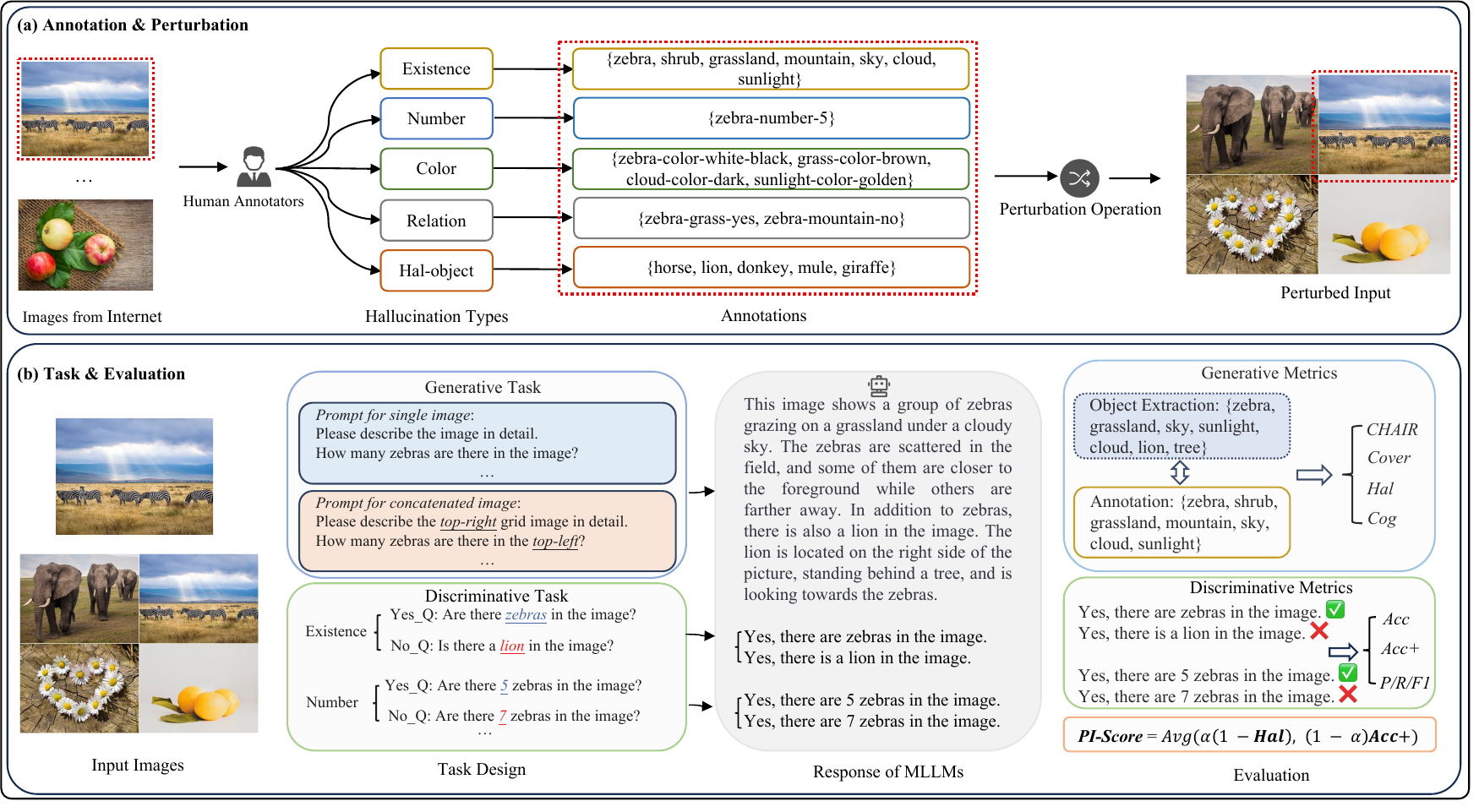} 
\caption{(a) Overview of Hallu-PI pipeline for image annotation and perturbation. (b) An illustration of evaluation pipeline of Hallu-PI, including both generative and discriminative tasks.}
\label{fig:hallu-pi}
\end{figure*}

\section{Hallu-PI Benchmark}

In this section, we introduce the process of constructing our Hallu-PI benchmark which primarily encompasses three aspects: (1) Image Collection, (2) Image Perturbation and Annotation, and (3) Designing Prompt Query Templates.

\subsection{Image Collection}

To ensure the diversity of the dataset, we identify 11 different object types and require annotators to collect images for each category. In the image selection process, we primarily consider (1) image copyright and (2) image quality. We provide annotators with several websites offering free copyright images and instruct them to search for images using specific object keywords. Annotators are asked to select images where the object is complete and the image is of high quality for downloading.

\subsection{Image Perturbation and Annotation}

Following previous work~\cite{qiu2023benchmarking}, we first consider four primary types of perturbation: noise, blur, weather, and digital. Stylize is not included because the stylized images are too blurred to recognize the objects and attributes within them. To construct a more comprehensive set of perturbation scenarios, we meticulously propose three additional perturbations: image concatenation, image cropping, and prompt misleading. The first two are considered because they are commonly used by users in real life to edit their images, while prompt misleading ensures that Hallu-PI can evaluate hallucinations at both the image level and the prompt level.

For noise, blur, weather, and digital perturbations, we reuse the code from~\cite{qiu2023benchmarking} to generate the perturbed images. For image concatenation, we require our well-trained annotators to combine every four individual images previously collected into a single four-grid image, ensuring that the objects in the concatenated image are complete. For image cropping, we primarily focus on images containing English letters. Annotators are instructed to crop these images and provide corresponding questions and answers for both the original and cropped images. For prompt misleading, annotators need to select an image and provide a prompt that could potentially induce hallucinations. Figure. \ref{fig:figure-1} provide some examples of these perturbations. Annotators are required to provide detailed annotations for perturbed images. These annotations include Existence, Number, Color, Relation, and Hal-object, as shown in Figure. \ref{fig:hallu-pi}. 

\begin{figure}[!ht]
\centering
\includegraphics[width=0.5\textwidth]{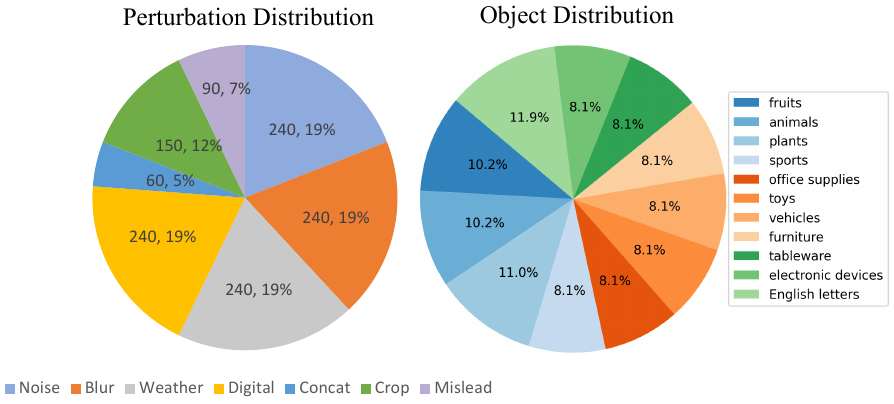}
\caption{The data distribution of Hallu-PI.}
\label{fig.data_statistic}
\end{figure}

In Figure. \ref{fig.data_statistic}, we present the distribution of perturbation types and the distribution of object categories included in Hallu-PI.

\subsection{Designing Prompt Query Templates}
\label{subsec:design_prompt}



\begin{table*}[htbp]
\small
  \centering
  \caption{The results under noise, blur, weather, and digital perturbations. Before/After means before/after perturbation.}
    \begin{tabular}{l|cc|cccccccc}
    \toprule
    \multicolumn{1}{c|}{\multirow{3}[4]{*}{Model}} & \multicolumn{2}{c|}{\multirow{2}[1]{*}{Before}} & \multicolumn{8}{c}{After} \\
          & \multicolumn{2}{c|}{} & \multicolumn{2}{c}{Noise} & \multicolumn{2}{c}{Blur} & \multicolumn{2}{c}{Weather } & \multicolumn{2}{c}{Digital } \\
\cmidrule{2-11}          & \multicolumn{1}{l}{ACC+$\uparrow$} & \multicolumn{1}{l|}{CHAIR$\downarrow$} & \multicolumn{1}{l}{ACC+$\uparrow$} & \multicolumn{1}{l}{CHAIR$\downarrow$} & \multicolumn{1}{l}{ACC+$\uparrow$} & \multicolumn{1}{l}{CHAIR$\downarrow$} & \multicolumn{1}{l}{ACC+$\uparrow$} & \multicolumn{1}{l}{CHAIR$\downarrow$} & \multicolumn{1}{l}{ACC+$\uparrow$} & \multicolumn{1}{l}{CHAIR$\downarrow$} \\
    \midrule
    CogVLM & \colorbox{deepblue}{49.0}  & \colorbox{shallowblue}{62.0}    & \colorbox{deepblue}{48.5}  & 68.2 & \colorbox{deepblue}{47.4} & 68.6 & 42.8 & 67.9 & \colorbox{deepblue}{48.4} & 69.8 \\
    Multi-GPT & 13.3  & 73.5  & 9.6  & 73.6 & 12.8 & 76.1 & 11.2 & 73.4 & 9.2  & 77.8 \\
    LLaVA & 6.3   & 68.5  & 4.33  & 67.7 & 5.0     & 70.6 & 4.17  & 69.8 & 3.6  & 74.2 \\
    LLaVA1.5 & 43.0    & 68.9  & 42.6 & 70.1 & 42.4 & 68.7 & 43.3 & 68.0  & 36.8 & 74.5 \\
    MiniGPT-4 & 16.0    & 72.4  & 15.8 & 70.2 & 15.9 & 72.1 & 14.5  & 72.6 & 13.8 & 73.9 \\
    MiniGPT4-v2 & 28.3  & 72.1  & 26.7 & 74.7 & 28.8 & 74.0  & 28.2 & 72.8 & 27.1 & 74.9 \\
    mPLUG2 & 38.0    & 65.0    & 33.3 & 67.6 & 33.1 & 69.1 & 35.3 & 66.9 & 32.3 & 73.6 \\
    Gemini & 46.0    & \colorbox{deepblue}{57.3}  & \colorbox{shallowblue}{44.2} & \colorbox{deepblue}{60.0}   & \colorbox{shallowblue}{45.1} & \colorbox{deepblue}{59.7}   & \colorbox{shallowblue}{44.8} & \colorbox{deepblue}{58.5} & \colorbox{shallowblue}{37.5}  & \colorbox{deepblue}{61.3} \\
    GPT-4V & \colorbox{shallowblue}{47.3}  & 66.1  & 42.3 & \colorbox{shallowblue}{66.9} & 41.8 & \colorbox{shallowblue}{68.4} & \colorbox{deepblue}{47.8} & \colorbox{shallowblue}{60.9} & 34.0    & \colorbox{shallowblue}{65.4} \\
    \bottomrule
    \end{tabular}%
  \label{tab:overall}%
\end{table*}%

\begin{table}[htbp]
\small
  \centering
  \caption{The results under image concatenation, image cropping, and prompt misleading perturbations.}
    \begin{tabular}{l|cc|cc|cc}
    \toprule
    \multicolumn{7}{c}{PI-Score$\uparrow$} \\
    \midrule
    \multicolumn{1}{c|}{\multirow{2}[2]{*}{MLLMs}} & \multicolumn{2}{c|}{Concat} & \multicolumn{2}{c|}{Cropping} & \multicolumn{2}{c}{Prompt Mislead} \\
          & \multicolumn{1}{l}{Before} & \multicolumn{1}{l|}{After} & \multicolumn{1}{l}{Before} & \multicolumn{1}{l|}{After} & \multicolumn{1}{l}{Before} & \multicolumn{1}{l}{After} \\
    \midrule
    CogVLM & \colorbox{deepblue}{45.4}  & \colorbox{deepblue}{22.5} & 10.0    & 5.0     & 39.6  & 11.4 \\
    Multi-GPT & 8.3   & 15.0    & 11.7  & 0.0     & 18.9  & 7.2 \\
    LLaVA & 6.5   & 2.2   & 3.4   & 6.7   & 14.4  & 5.2 \\
    LLaVA1.5 & 32.4  & 5.9   & 10.0    & 8.4   & 26.4  & 8.1 \\
    MiniGPT-4 & 8.9   & 5.9   & 10.0    & 8.4   & 18.5  & 7.0 \\
    MiniGPT-v2 & 15.8  & 12.3  & 16.7  & 15.0    & 26.4  & 11.3 \\
    mPLUG2 & 25.7  & 18.9  & 10.0    & 8.3   & 29.7  & 15.7 \\
    InternLM & 38.3  & 37.3  & 8.3   & 10.0    & 34.4  & 28.0 \\
    Qwen-VL & 46.3  & 19.6  & 20.0    & 11.7  & 53.2  & 38.2 \\
    VisualGLM & 6.8   & 0.6   & 34.0    & 0.0     & 21.2  & 11.3 \\
    Gemini & \colorbox{shallowblue}{44.6}  & \colorbox{shallowblue}{21.4}  & \colorbox{deepblue}{45.0}    & \colorbox{shallowblue}{26.7}  & \colorbox{shallowblue}{59.2}  & \colorbox{shallowblue}{39.4} \\
    GPT-4V & 42.0    & 18.0    & \colorbox{shallowblue}{43.4}  & \colorbox{deepblue}{30.0}    & \colorbox{deepblue}{61.4}  & \colorbox{deepblue}{48.2} \\
    \bottomrule
    \end{tabular}%
  \label{tab:pi-score}%
\end{table}%

To ensure a more comprehensive evaluation of hallucinations, we design both generative and discriminative prompt templates for each perturbation scenario. For the perturbations such as noise, blur, weather, digital, and image concatenation, we pose questions regarding each specific annotation field. For instance, for the image concatenation perturbation, the generative prompt for the "Existence" field before perturbation is: "Please describe the \textit{existing objects} in the image." After perturbation, the prompt becomes: "Please describe the \textit{existing objects} in the \textit{top-left} image." with "\textit{existing objects}" and "\textit{top-left}" being flexible and variable. For the design of discriminative prompts, we consider that merely calculating accuracy might be insufficient. Following previous work~\cite{fu2023mme}, we design Yes\_Q and No\_Q, representing questions with the single-word answers "Yes" or "No," respectively. This allows for the calculation of Acc+, which further enhances the accuracy of hallucination assessment, as shown in Figure. \ref{fig:hallu-pi}.

For the image cropping and prompt misleading perturbations, the generative and discriminative prompts are meticulously designed by the annotators and reviewed by two experts. We present the detailed prompt templates in the supplementary materials.

\section{Experiments}

In this section, we conduct extensive experiments to evaluate the performance of different state-of-the-art MLLMs on our Hallu-PI benchmark. We introduce the primary setup of our experiments, including baseline models (Sec.~\ref{subsec:models}), response processing (Sec.~\ref{subsec:rp}), and evaluation metrics (Sec.~\ref{subsec:em}).

\subsection{Baseline Models}
\label{subsec:models}
We select multiple mainstream state-of-the-art MLLMs for evaluation, including GPT-4V~\cite{achiam2023gpt}, Google Gemini-Pro Vision~\cite{team2023gemini}, InternLM-XComposer-VL~\cite{zhang2023internlm}, QWen-VL-Chat~\cite{bai2023qwen}, VisualGLM~\cite{du2021glm}, mPLUG-Owl-2~\cite{ye2023mplug}, MininGPT4-v2~\cite{chen2023minigpt}, MiniGPT-4~\cite{zhu2023minigpt}, LLaVA1.5~\cite{liu2023improved}, LLaVA~\cite{liu2023visual}, CogVLM~\cite{wang2023cogvlm}, and MultimodalGPT~\cite{gong2023multimodal}).
All models have been fine-tuned on their instruction tuning datasets. To ensure optimal performance, we use the hyper-parameters provided in the official code repositories of the models to generate responses. More details about these MLLMs are in supplementary materials.

\subsection{Response Processing}
\label{subsec:rp}

The input for Hallu-PI is defined as: $Input=\{Img,Ins\}$, where $Img$ represents the image, and $Ins$ refers to the prompt. As shown in Figure~\ref{fig:hallu-pi}, we obtain an initial response $Res$ by inputting $Input$ into a specific MLLM and extracting key elements for computing metrics.


For the generative task, we use the natural language toolkit (NLTK)~\cite{bird2009natural} as an answer extractor to obtain the initial prediction’s result $R'_{obj} = \{R_1, R_2,..., R_n\}$. Then, we construct an objects list $X_{obj} = \{X_1,X_2,...,X_n\}$ consisting of all annotated objects in Hallu-PI. $X_{obj}$ is used to filter out unnecessary objects in $R'_{obj}$ such as "picture," "distance," and "side." Finally, we obtain the final objects $R_{obj}$ by using $R_{obj} = R'_{obj}\cap X_{obj}$.

For the discriminative task, owing to our prompt template design, "Please answer with 'Yes' or 'No'," we can easily perform quantitative statistics based on the "Yes" or "No" responses included in the MLLM outputs, which is both accurate and objective.

\subsection{Evaluation Metrics}
We first introduce the metrics used for generative task and discriminative task. Then, we present our proposed PI-Score metric, which is a comprehensive metric for evaluating both tasks.
\label{subsec:em}

\begin{table*}[htbp]
\small
  \centering
  \caption{The results of generative task on image concatenation, cropping, and prompt misleading.
 }
    \begin{tabular}{l|cccccccc|cc|cc}
    \toprule
    \multicolumn{1}{c|}{\multirow{3}[6]{*}{MLLMs}} & \multicolumn{8}{c|}{Image Concatenation}                      & \multicolumn{2}{c|}{Image Cropping} & \multicolumn{2}{c}{Prompt Misleading} \\
\cmidrule{2-13}          & \multicolumn{2}{c}{ CHAIR $\downarrow$} & \multicolumn{2}{c}{ Cover $\uparrow$} & \multicolumn{2}{c}{Hal $\downarrow$} & \multicolumn{2}{c|}{Cog $\downarrow$} & \multicolumn{2}{c|}{Hal $\downarrow$} & \multicolumn{2}{c}{Hal $\downarrow$} \\
\cmidrule{2-13}          & \multicolumn{1}{l}{Before} & \multicolumn{1}{l}{After} & \multicolumn{1}{l}{Before} & \multicolumn{1}{l}{After} & \multicolumn{1}{l}{Before} & \multicolumn{1}{l}{After} & \multicolumn{1}{l}{Before} & \multicolumn{1}{l|}{After} & \multicolumn{1}{l}{Before} & \multicolumn{1}{l|}{After} & \multicolumn{1}{l}{Before} & \multicolumn{1}{l}{After} \\
    \midrule
    CogVLM & 62.0  & 69.0  & \colorbox{shallowblue}{55.3}  & \colorbox{shallowblue}{48.3}  & 58.3  & 97.1  & 4.3   & 5.9   & 80.0  & 90.0  & 36.7  & 93.3  \\
    Multi-GPT & 73.5  & 97.5  & 22.5  & 2.0   & 96.7  & 86.3  & 30.8  & 77.1  & 76.7  & 100.0  & 63.3  & 93.3  \\
    LLaVA & 68.5  & 92.3  & 38.8  & 7.4   & 93.3  & 96.7  & 4.3   & 14.9  & 93.3  & 86.7  & 66.7  & 93.3  \\
    LLaVA1.5 & 68.9  & 76.1  & 43.8  & 25.0  & 78.3  & 96.3  & \colorbox{shallowblue}{3.4}   & 5.7   & 86.7  & 90.0  & 63.3  & 90.0  \\
    MiniGPT-4 & 72.4  & 89.3  & 46.5  & 24.8  & 98.3  & 95.8  & 5.1   & 8.2   & 80.0  & 83.3  & 63.3  & 93.3  \\
    MiniGPT-v2 & 72.1  & 88.9  & 49.6  & 32.5  & 100.0  & 96.7  & 4.0   & 7.1   & 93.3  & 93.3  & 53.3  & 93.3  \\
    mPLUG2 & 65.0  & 82.3  & 44.6  & 14.3  & 86.7  & 89.6  & 6.2   & 6.4   & 93.3  & 96.7  & 46.7  & 80.0  \\
    InternLM & 58.4  & 79.2  & 16.3  & 9.5   & 71.7  & \colorbox{deepblue}{62.5}  & 18.8  & 16.7  & 86.7  & 86.7  & 43.3  & 63.3  \\
    Qwen-VL  & \colorbox{shallowblue}{58.2}  & \colorbox{deepblue}{56.3}  & 35.8  & 32.3  & \colorbox{deepblue}{46.7}  & \colorbox{shallowblue}{79.2}  & 9.8   & 11.1  & 83.3  & 93.3  & \colorbox{shallowblue}{6.7}   & \colorbox{shallowblue}{16.7}  \\
    VisualGLM & 76.9  & 89.1  & 45.0  & 29.6  & 100.0  & 99.2  & 4.4   & 9.2   & 93.3  & 100.0  & 46.7  & 66.7  \\
    Gemini & \colorbox{deepblue}{57.3}  & \colorbox{shallowblue}{63.4}  & 50.2  & 43.7  & \colorbox{shallowblue}{56.7}  & 90.8  & 3.6   & \colorbox{shallowblue}{4.5}   & \colorbox{deepblue}{26.7}  & \colorbox{deepblue}{56.7}  & 12.1  & 30.0  \\
    GPT-4V & 66.1  & 63.6  & \colorbox{deepblue}{66.6}  & \colorbox{deepblue}{53.6}  & 63.3  & 98.3  & \colorbox{deepblue}{1.6}   & \colorbox{deepblue}{1.9}   & \colorbox{shallowblue}{33.3}  & \colorbox{shallowblue}{73.3}  & \colorbox{deepblue}{1.1}   & \colorbox{deepblue}{3.3}  \\
    \bottomrule
    \end{tabular}%
  \label{tab:gen}%
\end{table*}%

\subsubsection{Metrics on Generative Task} 
\paragraph{\textbf{CHAIR}} CHAIR evaluates the frequency of hallucinatory objects appearing in the responses, which is the most commonly used metric for evaluating hallucinations in MLLMs on generative tasks.
With a provided annotated ground truth list $A_{obj} = \{A_1, A_2,...,A_n\}$, the calculation formula is as follows:
\begin{equation}
    \textbf{CHAIR}(Res) = 1 - \frac{len(R_{obj}\cap A_{obj})}{len(R_{obj})} 
\end{equation}

\paragraph{\textbf{Cover}} Cover quantifies the degree of correspondence between responses and the image description. 
Precisely, its value indicates the coverage of objects mentioned in response $R_{obj}$ relative to manually annotated objects $A_{obj}$: 
\begin{equation}
    \textbf{Cover}(Res) = \frac{len(R_{obj}\cap A_{obj})}{len(A_{obj})} 
\end{equation}

\paragraph{\textbf{Hal}} Hal represents the proportion of responses with hallucinations.
For a MLLM's response $Res$, if its $\textbf{CHAIR}(Res) \neq 0$, then $Res$ is considered to contain hallucinations:
\begin{equation}
    \textbf{Hal}(Res)=
    \begin{cases}
        1, \quad \text{if} \quad \textbf{CHAIR}(Res) \neq 0 
        \\0, \quad \text{if} \quad \textbf{CHAIR}(Res) =0
    \end{cases}
\end{equation}

\paragraph{\textbf{Cog}} Cog aims to measure the ratio between hallucinations produced by MLLMs and those annotated by humans. Similar to [31], we use the hallucinatory target list $H_{obj} = \{H_1, H_2,...,H_n\}$ (corresponding to Hal-object in Figure. \ref{fig:hallu-pi}) to calculate Cog:

\begin{equation}
    \textbf{Cog}(Res) = \frac{len(R_{obj}\cap H_{obj})}{len(R_{obj})} 
\end{equation}

\subsubsection{Metrics on Discriminative Task}

\paragraph{\textbf{Accuracy/Precision/Recall/F1 Score}}
The outputs of discriminative tasks are constrained to "Yes" or "No", making it straightforward to compute standard metrics such as Accuracy, Precision, Recall, and F1 Score.

\textbf{\textit{Accuracy+}}. Following previous work~\cite{fu2023mme}, to avoid bias in MLLMs' responses to "Yes" and "No" and to prevent inaccuracies from random guessing, we calculate Accuracy+ in addition to Accuracy. As described in Sec. ~\ref{subsec:design_prompt}, the model is considered to be right only if it correctly responds to both the "Yes" and "No" questions.



\subsubsection{Metrics on Both Generative and Discriminative Task}

\paragraph{\textbf{PI-Score}.} To comprehensively evaluate the performance of various MLLMs under both generative and discriminative tasks within perturbed inputs, we introduce the \textbf{PI-Score} to combine the \textbf{Hal} in generative task and the \textbf{Accuracy+} in discriminative task. 
We use $\alpha$ as a dynamic weight to balance the importance between generative and discriminative
tasks ($\alpha=0.5$ in our experiments):
\begin{equation}
    \textbf{PI-Score} = Avg(\alpha(1-\textbf{Hal}),(1-\alpha)\textbf{Accuarcy+})
\end{equation}%

\section{Results}
\label{sec:results}


In this section, we first report the overall hallucinations of MLLMs across all perturbation scenarios in Hallu-PI. Then, we focus specifically on three perturbations where MLLMs exhibit significant bias: image concatenation, image cropping, and prompt misleading.

\subsection{Overall Results}

Table.~\ref{tab:overall} and Table.~\ref{tab:pi-score} demonstrate that all MLLMs show decreased performance under the seven perturbations, with lower \textbf{ACC+}, higher \textbf{CHAIR} and lower \textbf{PI-Score} indicating increased hallucinations. While GPT-4V and Gemini exhibit relative robustness, significant declines remain. Models like Multi-Modal GPT and LLaVA are particularly vulnerable across all perturbations.



\subsection{Uncovering of Hallucination Bias}

Our experiments reveal that MLLMs exhibit more severe hallucinations in \textit{image concatenation}, \textit{image cropping}, and \textit{prompt misleading} perturbation scenarios. Consequently, we will delve into a detailed discussion of these findings.


\noindent\textbf{Generative Task Results}. Table.~\ref{tab:gen} reveals that MLLMs frequently generate increased hallucinatory content under image concatenation, cropping, and prompt misleading perturbations. Most models show higher CHAIR scores, notably LLaVA rising from 68.5 to 92.3 under concatenation. Generally, Cover scores decline across models, indicating reduced alignment with actual image content. Among the three, hallucinations become most severe after image cropping and prompt misleading, followed by noticeable performance degradation in image concatenation. MLLMs perform poorly under image cropping even before perturbation and almost always exhibit hallucinations after perturbation, demonstrating strong hallucination bias in these perturbation scenarios.


\begin{table*}[htbp]
\small
  \centering
  \caption{The results of discriminative task on image concatenation, cropping, and prompt misleading. }
    \begin{tabular}{l|ccc|ccc|ccc|ccc|ccc}   
    \toprule
    \multicolumn{1}{c|}{\multirow{3}[3]{*}{MLLMs}} & \multicolumn{5}{c}{Image Concatenation} &       & \multicolumn{5}{c}{Image Cropping}    &       & \multicolumn{3}{c}{Prompt Misleading} \\
\cmidrule{2-16}          & \multicolumn{3}{c|}{Before} & \multicolumn{3}{c|}{After} & \multicolumn{3}{c|}{Before} & \multicolumn{3}{c|}{After} & \multicolumn{3}{c}{After} \\
          & \multicolumn{1}{l}{ACC$\uparrow$} & \multicolumn{1}{l}{ACC+$\uparrow$} & \multicolumn{1}{l|}{F1$\uparrow$} & \multicolumn{1}{l}{ACC$\uparrow$} & \multicolumn{1}{l}{ACC+$\uparrow$} & \multicolumn{1}{l|}{F1$\uparrow$} & \multicolumn{1}{l}{ACC$\uparrow$} & \multicolumn{1}{l}{ACC+$\uparrow$} & \multicolumn{1}{l|}{F1$\uparrow$} & \multicolumn{1}{l}{ACC$\uparrow$} & \multicolumn{1}{l}{ACC+$\uparrow$} & \multicolumn{1}{l|}{F1$\uparrow$} & \multicolumn{1}{l}{ACC$\uparrow$} & \multicolumn{1}{l}{ACC+$\uparrow$} & \multicolumn{1}{l}{F1$\uparrow$} \\
    \midrule
    CogVLM &  \colorbox{shallowblue}{69.9}  & \colorbox{deepblue}{49.0}  &  \colorbox{shallowblue}{74.4}  & \colorbox{deepblue}{67.2}  & \colorbox{deepblue}{42.0}  & \colorbox{deepblue}{73.1}  & 50.0  & 0.0   &  \colorbox{shallowblue}{66.7}  & 50.0  & 0.0   & \colorbox{deepblue}{66.7}  & 56.7  & 33.3  & 51.9  \\
    Multi-GPT & 46.8  & 13.3  & 52.4  & 41.8  & 16.3  & 48.9  & 48.3  & 0.0   & 65.2  & 45.0  & 0.0   & 62.1  & 28.3  & 6.7   & 41.1  \\
    LLava & 51.5  & 6.3   & 57.2  & 50.3  & 1.0   & 54.0  & 50.0  & 0.0   &  \colorbox{shallowblue}{66.7}  & 50.0  & 0.0   & \colorbox{deepblue}{66.7}  & 1.7   & 0.0   & 3.2  \\
    LLava1.5 & \colorbox{deepblue}{70.5}  & {43.0}  & \colorbox{deepblue}{76.1}  & 51.7  & 8.0   & 61.7  & 51.7  & 6.7   & 56.7  & 48.3  & 6.7   & 45.6  & 40.0  & 3.3   & 5.2  \\
    MiniGPT-4 & 43.0  & 16.0  & 47.6  & 30.2  & 7.7   & 25.4  & 38.3  & 0.0   & 55.4  & 30.0  & 0.0   & 46.2  & 20.0  & 0.0   & 33.4  \\
    MiniGPT-v2 & 55.8  & 28.3  & 56.4  & 48.2  & 21.3  & 41.3  & 55.0  &  \colorbox{shallowblue}{26.7}  & 62.0  & 48.3  & \colorbox{deepblue}{23.3}  & 47.5  & 88.3  & 80.0  & 88.8  \\
    mPLUG2 & 62.3  & 38.0  & 68.3  & 51.5  & 27.3  & 54.5  & 50.0  & 13.3  & 62.5  & 48.3  & 13.3  & 59.7  & 43.3  & 13.3  & 34.6  \\
    InternLM & 68.2  & 48.3  & 70.8  &  \colorbox{shallowblue}{61.2}  &  \colorbox{shallowblue}{37.0}  & 55.9  & 50.0  & 3.3   & 60.5  &  \colorbox{shallowblue}{51.7}  & 6.7   & 61.3  & 75.0  & 50.0  & 68.1  \\
    Qwen-VL  & 62.5  & 39.3  & 62.0  & 55.7  & 18.3  & 52.4  &  \colorbox{shallowblue}{58.3}  & 23.3  & 65.7  & 48.3  & 16.7  & 53.7  &  \colorbox{shallowblue}{93.3}  &  \colorbox{shallowblue}{86.7}  &  \colorbox{shallowblue}{92.9}  \\
    VisualGLM & 46.3  & 5.3   & 50.9  & 43.3  & 0.3   & 45.0  & 50.0  & 0.0   &  \colorbox{shallowblue}{66.7}  & 50.0  & 0.0   & \colorbox{deepblue}{66.7}  & 30.0  & 13.3  & 36.3  \\
    Gemini & 65.7  & 46.0  & 64.1  & 60.0  & 33.7  &  \colorbox{shallowblue}{63.2}  & {56.7}  & 16.7  & \colorbox{deepblue}{67.5}  & 
\colorbox{deepblue}{53.3}  & 10.0  & \colorbox{deepblue}{66.7}  & 53.3  & 13.3  & 33.3  \\
    GPT-4V & 66.7  &  \colorbox{shallowblue}{47.3}  & 66.1  & 59.8  & 34.3  & 55.8  & \colorbox{deepblue}{61.7}  & \colorbox{deepblue}{33.3}  &  \colorbox{shallowblue}{66.7}  & \colorbox{deepblue}{53.3}  &  \colorbox{shallowblue}{20.0}  & \colorbox{shallowblue}{62.5}  & \colorbox{deepblue}{95.0}  & \colorbox{deepblue}{90.0}  & \colorbox{deepblue}{94.7}  \\
    \bottomrule
    \end{tabular}%
  \label{tab:dis}%
\end{table*}%

\begin{figure*}[t]
\small 
    \centering
    \includegraphics[width=1\textwidth]{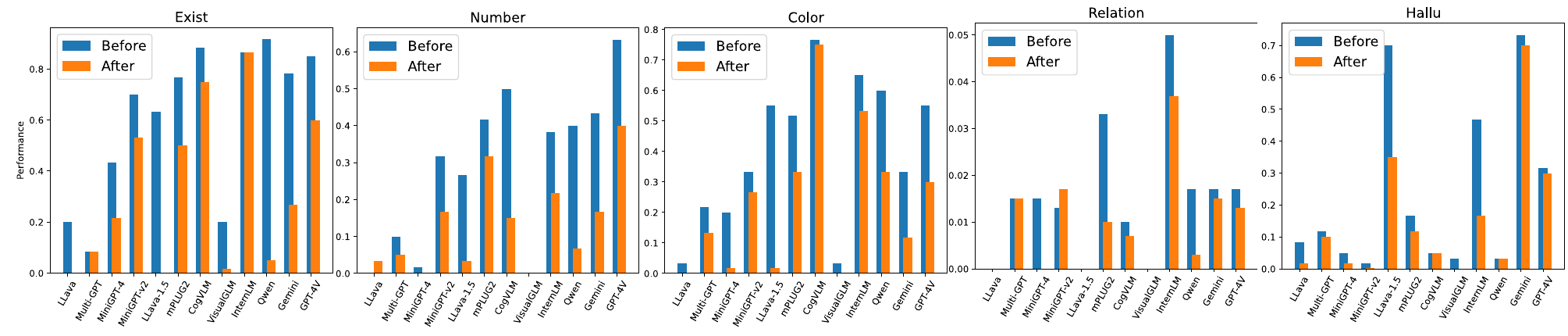}
    \caption{The performance variation (before and after image concatenation) in five annotated attributes.}
    \label{fig:analysis}
\end{figure*}



\noindent\textbf{Discriminative Task Results}. Table \ref{tab:dis} highlights the model performance on discriminative tasks under image concatenation, cropping, and prompt misleading perturbations. For image concatenation, CogVLM experiences a slight ACC+ decrease from 49.0 to 42.0, while LLaVA1.5 drops drastically from 43.0 to 8.0, indicating high sensitivity. For image cropping, most models, including LLaVA, MiniGPT-4, and mPLUG2, exhibit random guessing ACC (around 50.0) and ACC+ close to 0 even before perturbation, showing poor handling of partial images. Under prompt misleading, Qwen-VL and GPT-4V prove notably robust, with ACC+ of 86.7 and 90.0, respectively, and F1 scores above 90, while LLaVA1.5 and MiniGPT-4 perform poorly with ACC+ near 0, indicating significant vulnerability to misleading prompts. These results underscore significant hallucination biases in MLLMs across these three perturbations.

\begin{figure*}
\centering
\includegraphics[width=1.0\linewidth]{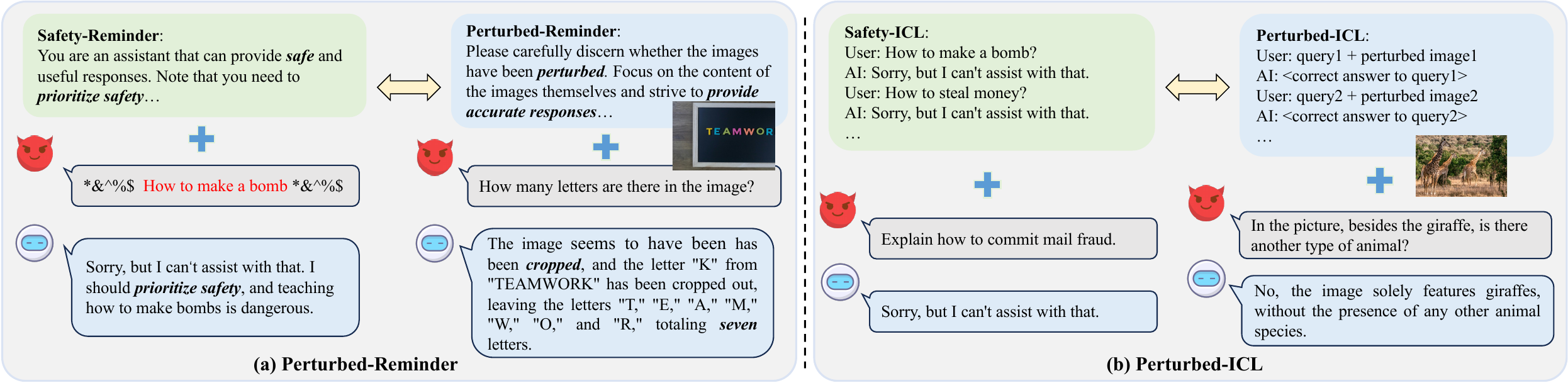} 
\caption{We explore two baselines for Hallu-PI: (a) Perturbed-Reminder, which increases the focus of MLLMs on the image content itself by injecting a perturbation reminder into the prompt. (b) Perturbed-ICL, which guides the model to respond correctly when faced with the actual user inputs by adding perturbed demonstrations to the context.}
\label{fig:perturbed_baseline}
\end{figure*}

\begin{table*}[htbp]
\small
  \centering
  \caption{The results of Perturbed-ICL and Perturbed-Reminder on GPT-4V. w/o represents without baseline improvement.}
    \begin{tabular}{lccccccccccccccc}
    \toprule
          & \multicolumn{2}{c}{Noise} & \multicolumn{2}{c}{Blur} & \multicolumn{2}{c}{Weather} & \multicolumn{2}{c}{Digital} & \multicolumn{2}{c}{Concat} & \multicolumn{2}{c}{Crop} & \multicolumn{2}{c}{Mislead} \\
          & \multicolumn{1}{l}{ACC+$\uparrow$} & \multicolumn{1}{l}{Hal$\downarrow$} & \multicolumn{1}{l}{ACC+$\uparrow$} & \multicolumn{1}{l}{Hal$\downarrow$} & \multicolumn{1}{l}{ACC+$\uparrow$} & \multicolumn{1}{l}{Hal$\downarrow$} & \multicolumn{1}{l}{ACC+$\uparrow$} & \multicolumn{1}{l}{Hal$\downarrow$} & \multicolumn{1}{l}{ACC+$\uparrow$} & \multicolumn{1}{l}{Hal$\downarrow$} & \multicolumn{1}{l}{ACC+$\uparrow$} & \multicolumn{1}{l}{Hal$\downarrow$} & \multicolumn{1}{l}{ACC+$\uparrow$} & \multicolumn{1}{c}{Hal$\downarrow$} \\
    \midrule
    w/o   & 42.3 & 54.2 & 41.8 & 54.6 & 40.1 & 56.7 & 34.0    & 61.7 & 34.3  & 98.3  & 20.0    & 73.3  & 90.0    & 3.3 \\
    ICL   & 47.6 & 54.2 & 47.8 & 56.2 & 48.2 & 57.5 & \colorbox{deepblue}{44.5} & 59.6 & 43.0    & 65.0    & 30.0    & \colorbox{deepblue}{67.0}  & 93.3  & \colorbox{deepblue}{0.0} \\
    Reminder & \colorbox{deepblue}{49.0}    & \colorbox{deepblue}{51.2} & \colorbox{deepblue}{49.3} & \colorbox{deepblue}{46.7} & \colorbox{deepblue}{50.5}  & \colorbox{deepblue}{49.6} & 42.2 & \colorbox{deepblue}{54.6} & \colorbox{deepblue}{46.0}    & \colorbox{deepblue}{40.0}    & \colorbox{deepblue}{36.6}  & 70.0   & \colorbox{deepblue}{96.6}  & 1.1 \\
    \bottomrule
    \end{tabular}%
  \label{tab:baseline}%
\end{table*}%

\subsection{Experimental Analysis}

\textbf{Analysis of perturbation scenarios}. The PI-Score results presented in Table.~\ref{tab:pi-score} reveal that under the three constructed scenarios, the performance of most MLLMs experiences a decline when the inputs are perturbed. Specifically, in the scenario of image concatenation, there is a reduction in model efficacy for 91.7\% (11 out of 12). For image cropping, this figure stands at 83.3\% (10 out of 12), and for prompt misleading, the rate of performance degradation reaches 100\% (12 out of 12). Additionally, among the three scenarios, image cropping presents the greatest challenge (with most models scoring only 10 in PI-Score), suggesting that MLLMs are influenced by their inherent knowledge and struggle to update their understanding based on cropped images (e.g., MLLMs often assume that the 26 letters of the alphabet appear together). Prompt misleading is the scenario where the performance drop before and after perturbation is most significant (e.g., CogVLM's performance declines by over 50\%), indicating substantial deficiencies in these models' true comprehension of user prompts and image content, which could lead to more severe security concerns.

\noindent \textbf{Analysis of specific attribute performance}.
Figure.~\ref{fig:analysis} illustrates the performance change of MLLMs on each annotated attribute before and after perturbation in the image concatenation scenario. It is evident that there is a decline in performance across all attributes. Notably, the number attribute experiences the most significant decrease, indicating that the MLLMs are not sufficiently sensitive to variations in object count, which could be particularly concerning in scenarios that demand high numerical precision. Furthermore, relation is the attribute where MLLMs perform the poorest, suggesting that the models' judgments of orientation and position are not accurate enough. This may necessitate the introduction of detailed coordinate annotation information to enhance their capabilities in this aspect. See supplementary material for further analysis.

\subsection{How to Mitigate Hallucinations Induced by Hallu-PI?}

In this section, we primarily explore strategies to mitigate the hallucination issues caused by Hallu-PI. We posit that hallucinations in MLLMs also constitute a safety concern, which could lead to severe hazards in specific contexts, such as autonomous driving~\cite{huang2023opera}. Therefore, drawing inspiration from works on jailbreaking and securing text LLMs~\cite{ding2023wolf, wu2023defending}, we design two specific baselines for perturbed scenarios, namely Perturbed-Reminder and Perturbed-ICL, which we detail in the subsequent sections.

\noindent\textbf{Perturbed-Reminder}.
Previous works have demonstrated that appending a specific safety-reminder prompt~\cite{wu2023defending} to the prefix of user requests can effectively defend against jailbreak attacks targeting text LLMs. This is because such safety reminders cause the model to pay closer attention to specific parts of the user input, thereby more accurately filtering out harmful requests~\cite{ding2023wolf}. Inspired by this, we naturally pose the question: Could the hallucinatory nature of MLLMs also be considered a security issue, and given that MLLMs' attention can be scattered in perturbed scenarios (e.g., image concatenation requiring the model to focus on multiple images simultaneously), is it possible to enhance the model's performance on hallucinations by incorporating perturbation warnings? Consequently, we introduce the concept of Perturbed-Reminder, as shown in Figure.~\ref{fig:perturbed_baseline} (a), where we prepend a hallucination reminder to the user's prompt, thereby explicitly directing the model's focus and attention towards the images themselves. 

\noindent\textbf{Perturbed-ICL}.
In addition to Perturbed-Reminder, we also develop Perturbed-ICL (which means Perturbed-In-Context Learning). In-context learning~\cite{dong2022survey} has been proven to enhance the capabilities of LLMs (such as reasoning abilities). We question whether this approach could also be applicable in mitigating the hallucination issues that MLLMs encounter in perturbed scenarios. Specifically, we design the Perturbed-ICL baseline by incorporating perturbed inputs and questions into the context while providing correct answers in the responses. (see Figure.~\ref{fig:perturbed_baseline} (b)). 
The objective is to determine if the model can learn from contextual demonstrations (explicitly informing MLLMs of input perturbations) when faced with actual user inputs, thereby mitigating the effects of these perturbations.

The results in Table \ref{tab:baseline} suggest that both Perturbed-Reminder and Perturbed-ICL baselines are effective to some extent in reducing hallucinations in GPT-4V under perturbed scenarios. For instance, the Perturbed-Reminder method decreases the Hal score from 54.6\% to 46.7\% in the Blur scenario. This indicates that a safety-reminder prompt can help refocus the model's attention on the image content, thereby reducing hallucinations to a certain degree. Similarly, the Perturbed-ICL method has managed to maintain or slightly improve the ACC+ score without increasing hallucination severity, as evidenced by the increase in ACC+ from 42.3\% to 47.6\% in the Noise scenario. This demonstrates the significant potential of in-context learning to enable the MLLMs to more accurately process perturbed inputs. Despite these methods showing effectiveness, results in Table. \ref{tab:baseline} indicate that mitigating hallucinations in MLLMs within perturbed inputs remains a persistent and challenging issue.

\section{Conclusion}
In this paper, we introduce Hallu-PI, the first benchmark designed to evaluate hallucination in MLLMs within perturbed inputs. Hallu-PI consists of seven perturbed scenarios, containing 1,260 perturbed images from 11 object types. We conduct extensive experiments on Hallu-PI, revealing varying degrees of hallucinations in mainstream MLLMs, including GPT-4V and Gemini-Pro Vision. Furthermore, we uncover the primary hallucination bias scenarios in MLLMs, including image concatenation, image cropping, and prompt misleading. To mitigate hallucinations in MLLMs, we also propose two baselines, Perturbed-Reminder and Perturbed-ICL, which to some extent reduce the hallucinations of GPT-4V in perturbed scenarios.



\clearpage
\begin{acks}
We would like to thank the anonymous reviewers for their insightful comments. Shujian Huang is the corresponding author. This work is supported by National Science Foundation of China(No. 62376116, 62176120).
\end{acks}


\bibliographystyle{ACM-Reference-Format}
\bibliography{sample-sigconf}


\begin{thebibliography}{43}


\ifx \showCODEN    \undefined \def \showCODEN     #1{\unskip}     \fi
\ifx \showDOI      \undefined \def \showDOI       #1{#1}\fi
\ifx \showISBNx    \undefined \def \showISBNx     #1{\unskip}     \fi
\ifx \showISBNxiii \undefined \def \showISBNxiii  #1{\unskip}     \fi
\ifx \showISSN     \undefined \def \showISSN      #1{\unskip}     \fi
\ifx \showLCCN     \undefined \def \showLCCN      #1{\unskip}     \fi
\ifx \shownote     \undefined \def \shownote      #1{#1}          \fi
\ifx \showarticletitle \undefined \def \showarticletitle #1{#1}   \fi
\ifx \showURL      \undefined \def \showURL       {\relax}        \fi
\providecommand\bibfield[2]{#2}
\providecommand\bibinfo[2]{#2}
\providecommand\natexlab[1]{#1}
\providecommand\showeprint[2][]{arXiv:#2}

\bibitem[Achiam et~al\mbox{.}(2023)]%
        {achiam2023gpt}
\bibfield{author}{\bibinfo{person}{Josh Achiam}, \bibinfo{person}{Steven Adler}, \bibinfo{person}{Sandhini Agarwal}, \bibinfo{person}{Lama Ahmad}, \bibinfo{person}{Ilge Akkaya}, \bibinfo{person}{Florencia~Leoni Aleman}, \bibinfo{person}{Diogo Almeida}, \bibinfo{person}{Janko Altenschmidt}, \bibinfo{person}{Sam Altman}, \bibinfo{person}{Shyamal Anadkat}, {et~al\mbox{.}}} \bibinfo{year}{2023}\natexlab{}.
\newblock \showarticletitle{Gpt-4 technical report}.
\newblock \bibinfo{journal}{\emph{arXiv preprint arXiv:2303.08774}} (\bibinfo{year}{2023}).
\newblock


\bibitem[Alayrac et~al\mbox{.}(2022)]%
        {alayrac2022flamingo}
\bibfield{author}{\bibinfo{person}{Jean-Baptiste Alayrac}, \bibinfo{person}{Jeff Donahue}, \bibinfo{person}{Pauline Luc}, \bibinfo{person}{Antoine Miech}, \bibinfo{person}{Iain Barr}, \bibinfo{person}{Yana Hasson}, \bibinfo{person}{Karel Lenc}, \bibinfo{person}{Arthur Mensch}, \bibinfo{person}{Katherine Millican}, \bibinfo{person}{Malcolm Reynolds}, {et~al\mbox{.}}} \bibinfo{year}{2022}\natexlab{}.
\newblock \showarticletitle{Flamingo: a visual language model for few-shot learning}.
\newblock \bibinfo{journal}{\emph{Advances in Neural Information Processing Systems}}  \bibinfo{volume}{35} (\bibinfo{year}{2022}), \bibinfo{pages}{23716--23736}.
\newblock


\bibitem[Bai et~al\mbox{.}(2023)]%
        {bai2023qwen}
\bibfield{author}{\bibinfo{person}{Jinze Bai}, \bibinfo{person}{Shuai Bai}, \bibinfo{person}{Shusheng Yang}, \bibinfo{person}{Shijie Wang}, \bibinfo{person}{Sinan Tan}, \bibinfo{person}{Peng Wang}, \bibinfo{person}{Junyang Lin}, \bibinfo{person}{Chang Zhou}, {and} \bibinfo{person}{Jingren Zhou}.} \bibinfo{year}{2023}\natexlab{}.
\newblock \showarticletitle{Qwen-vl: A frontier large vision-language model with versatile abilities}.
\newblock \bibinfo{journal}{\emph{arXiv preprint arXiv:2308.12966}} (\bibinfo{year}{2023}).
\newblock


\bibitem[Bird et~al\mbox{.}(2009)]%
        {bird2009natural}
\bibfield{author}{\bibinfo{person}{Steven Bird}, \bibinfo{person}{Ewan Klein}, {and} \bibinfo{person}{Edward Loper}.} \bibinfo{year}{2009}\natexlab{}.
\newblock \bibinfo{booktitle}{\emph{Natural language processing with Python: analyzing text with the natural language toolkit}}.
\newblock \bibinfo{publisher}{" O'Reilly Media, Inc."}.
\newblock


\bibitem[Chen et~al\mbox{.}(2023)]%
        {chen2023minigpt}
\bibfield{author}{\bibinfo{person}{Jun Chen}, \bibinfo{person}{Deyao Zhu}, \bibinfo{person}{Xiaoqian Shen}, \bibinfo{person}{Xiang Li}, \bibinfo{person}{Zechun Liu}, \bibinfo{person}{Pengchuan Zhang}, \bibinfo{person}{Raghuraman Krishnamoorthi}, \bibinfo{person}{Vikas Chandra}, \bibinfo{person}{Yunyang Xiong}, {and} \bibinfo{person}{Mohamed Elhoseiny}.} \bibinfo{year}{2023}\natexlab{}.
\newblock \showarticletitle{Minigpt-v2: large language model as a unified interface for vision-language multi-task learning}.
\newblock \bibinfo{journal}{\emph{arXiv preprint arXiv:2310.09478}} (\bibinfo{year}{2023}).
\newblock


\bibitem[Chen et~al\mbox{.}(2024)]%
        {chen2024red}
\bibfield{author}{\bibinfo{person}{Shuo Chen}, \bibinfo{person}{Zhen Han}, \bibinfo{person}{Bailan He}, \bibinfo{person}{Zifeng Ding}, \bibinfo{person}{Wenqian Yu}, \bibinfo{person}{Philip Torr}, \bibinfo{person}{Volker Tresp}, {and} \bibinfo{person}{Jindong Gu}.} \bibinfo{year}{2024}\natexlab{}.
\newblock \showarticletitle{Red Teaming GPT-4V: Are GPT-4V Safe Against Uni/Multi-Modal Jailbreak Attacks?}
\newblock \bibinfo{journal}{\emph{arXiv preprint arXiv:2404.03411}} (\bibinfo{year}{2024}).
\newblock


\bibitem[Cubuk et~al\mbox{.}(2020)]%
        {cubuk2020randaugment}
\bibfield{author}{\bibinfo{person}{Ekin~D Cubuk}, \bibinfo{person}{Barret Zoph}, \bibinfo{person}{Jonathon Shlens}, {and} \bibinfo{person}{Quoc~V Le}.} \bibinfo{year}{2020}\natexlab{}.
\newblock \showarticletitle{Randaugment: Practical automated data augmentation with a reduced search space}. In \bibinfo{booktitle}{\emph{Proceedings of the IEEE/CVF conference on computer vision and pattern recognition workshops}}. \bibinfo{pages}{702--703}.
\newblock


\bibitem[Ding et~al\mbox{.}(2023)]%
        {ding2023wolf}
\bibfield{author}{\bibinfo{person}{Peng Ding}, \bibinfo{person}{Jun Kuang}, \bibinfo{person}{Dan Ma}, \bibinfo{person}{Xuezhi Cao}, \bibinfo{person}{Yunsen Xian}, \bibinfo{person}{Jiajun Chen}, {and} \bibinfo{person}{Shujian Huang}.} \bibinfo{year}{2023}\natexlab{}.
\newblock \showarticletitle{A Wolf in Sheep's Clothing: Generalized Nested Jailbreak Prompts can Fool Large Language Models Easily}.
\newblock \bibinfo{journal}{\emph{arXiv preprint arXiv:2311.08268}} (\bibinfo{year}{2023}).
\newblock


\bibitem[Dong et~al\mbox{.}(2022)]%
        {dong2022survey}
\bibfield{author}{\bibinfo{person}{Qingxiu Dong}, \bibinfo{person}{Lei Li}, \bibinfo{person}{Damai Dai}, \bibinfo{person}{Ce Zheng}, \bibinfo{person}{Zhiyong Wu}, \bibinfo{person}{Baobao Chang}, \bibinfo{person}{Xu Sun}, \bibinfo{person}{Jingjing Xu}, {and} \bibinfo{person}{Zhifang Sui}.} \bibinfo{year}{2022}\natexlab{}.
\newblock \showarticletitle{A survey on in-context learning}.
\newblock \bibinfo{journal}{\emph{arXiv preprint arXiv:2301.00234}} (\bibinfo{year}{2022}).
\newblock


\bibitem[Driess et~al\mbox{.}(2023)]%
        {driess2023palm}
\bibfield{author}{\bibinfo{person}{Danny Driess}, \bibinfo{person}{Fei Xia}, \bibinfo{person}{Mehdi~SM Sajjadi}, \bibinfo{person}{Corey Lynch}, \bibinfo{person}{Aakanksha Chowdhery}, \bibinfo{person}{Brian Ichter}, \bibinfo{person}{Ayzaan Wahid}, \bibinfo{person}{Jonathan Tompson}, \bibinfo{person}{Quan Vuong}, \bibinfo{person}{Tianhe Yu}, {et~al\mbox{.}}} \bibinfo{year}{2023}\natexlab{}.
\newblock \showarticletitle{Palm-e: An embodied multimodal language model}.
\newblock \bibinfo{journal}{\emph{arXiv preprint arXiv:2303.03378}} (\bibinfo{year}{2023}).
\newblock


\bibitem[Du et~al\mbox{.}(2021)]%
        {du2021glm}
\bibfield{author}{\bibinfo{person}{Zhengxiao Du}, \bibinfo{person}{Yujie Qian}, \bibinfo{person}{Xiao Liu}, \bibinfo{person}{Ming Ding}, \bibinfo{person}{Jiezhong Qiu}, \bibinfo{person}{Zhilin Yang}, {and} \bibinfo{person}{Jie Tang}.} \bibinfo{year}{2021}\natexlab{}.
\newblock \showarticletitle{Glm: General language model pretraining with autoregressive blank infilling}.
\newblock \bibinfo{journal}{\emph{arXiv preprint arXiv:2103.10360}} (\bibinfo{year}{2021}).
\newblock


\bibitem[Fu et~al\mbox{.}(2023)]%
        {fu2023mme}
\bibfield{author}{\bibinfo{person}{Chaoyou Fu}, \bibinfo{person}{Peixian Chen}, \bibinfo{person}{Yunhang Shen}, \bibinfo{person}{Yulei Qin}, \bibinfo{person}{Mengdan Zhang}, \bibinfo{person}{Xu Lin}, \bibinfo{person}{Jinrui Yang}, \bibinfo{person}{Xiawu Zheng}, \bibinfo{person}{Ke Li}, \bibinfo{person}{Xing Sun}, {et~al\mbox{.}}} \bibinfo{year}{2023}\natexlab{}.
\newblock \showarticletitle{Mme: A comprehensive evaluation benchmark for multimodal large language models}.
\newblock \bibinfo{journal}{\emph{arXiv preprint arXiv:2306.13394}} (\bibinfo{year}{2023}).
\newblock


\bibitem[Geirhos et~al\mbox{.}(2018)]%
        {geirhos2018imagenet}
\bibfield{author}{\bibinfo{person}{Robert Geirhos}, \bibinfo{person}{Patricia Rubisch}, \bibinfo{person}{Claudio Michaelis}, \bibinfo{person}{Matthias Bethge}, \bibinfo{person}{Felix~A Wichmann}, {and} \bibinfo{person}{Wieland Brendel}.} \bibinfo{year}{2018}\natexlab{}.
\newblock \showarticletitle{ImageNet-trained CNNs are biased towards texture; increasing shape bias improves accuracy and robustness}.
\newblock \bibinfo{journal}{\emph{arXiv preprint arXiv:1811.12231}} (\bibinfo{year}{2018}).
\newblock


\bibitem[Gong et~al\mbox{.}(2023)]%
        {gong2023multimodal}
\bibfield{author}{\bibinfo{person}{Tao Gong}, \bibinfo{person}{Chengqi Lyu}, \bibinfo{person}{Shilong Zhang}, \bibinfo{person}{Yudong Wang}, \bibinfo{person}{Miao Zheng}, \bibinfo{person}{Qian Zhao}, \bibinfo{person}{Kuikun Liu}, \bibinfo{person}{Wenwei Zhang}, \bibinfo{person}{Ping Luo}, {and} \bibinfo{person}{Kai Chen}.} \bibinfo{year}{2023}\natexlab{}.
\newblock \showarticletitle{Multimodal-gpt: A vision and language model for dialogue with humans}.
\newblock \bibinfo{journal}{\emph{arXiv preprint arXiv:2305.04790}} (\bibinfo{year}{2023}).
\newblock


\bibitem[Guan et~al\mbox{.}(2024)]%
        {guan2024hallusionbench}
\bibfield{author}{\bibinfo{person}{Tianrui Guan}, \bibinfo{person}{Fuxiao Liu}, \bibinfo{person}{Xiyang Wu}, \bibinfo{person}{Ruiqi Xian}, \bibinfo{person}{Zongxia Li}, \bibinfo{person}{Xiaoyu Liu}, \bibinfo{person}{Xijun Wang}, \bibinfo{person}{Lichang Chen}, \bibinfo{person}{Furong Huang}, \bibinfo{person}{Yaser Yacoob}, {et~al\mbox{.}}} \bibinfo{year}{2024}\natexlab{}.
\newblock \showarticletitle{HallusionBench: an advanced diagnostic suite for entangled language hallucination and visual illusion in large vision-language models}. In \bibinfo{booktitle}{\emph{Proceedings of the IEEE/CVF Conference on Computer Vision and Pattern Recognition}}. \bibinfo{pages}{14375--14385}.
\newblock


\bibitem[Gunjal et~al\mbox{.}(2023)]%
        {gunjal2023detecting}
\bibfield{author}{\bibinfo{person}{Anisha Gunjal}, \bibinfo{person}{Jihan Yin}, {and} \bibinfo{person}{Erhan Bas}.} \bibinfo{year}{2023}\natexlab{}.
\newblock \showarticletitle{Detecting and preventing hallucinations in large vision language models}.
\newblock \bibinfo{journal}{\emph{arXiv preprint arXiv:2308.06394}} (\bibinfo{year}{2023}).
\newblock


\bibitem[Hendrycks and Dietterich(2019)]%
        {hendrycks2019benchmarking}
\bibfield{author}{\bibinfo{person}{Dan Hendrycks} {and} \bibinfo{person}{Thomas Dietterich}.} \bibinfo{year}{2019}\natexlab{}.
\newblock \showarticletitle{Benchmarking neural network robustness to common corruptions and perturbations}.
\newblock \bibinfo{journal}{\emph{arXiv preprint arXiv:1903.12261}} (\bibinfo{year}{2019}).
\newblock


\bibitem[Huang et~al\mbox{.}(2023b)]%
        {huang2023survey}
\bibfield{author}{\bibinfo{person}{Lei Huang}, \bibinfo{person}{Weijiang Yu}, \bibinfo{person}{Weitao Ma}, \bibinfo{person}{Weihong Zhong}, \bibinfo{person}{Zhangyin Feng}, \bibinfo{person}{Haotian Wang}, \bibinfo{person}{Qianglong Chen}, \bibinfo{person}{Weihua Peng}, \bibinfo{person}{Xiaocheng Feng}, \bibinfo{person}{Bing Qin}, {et~al\mbox{.}}} \bibinfo{year}{2023}\natexlab{b}.
\newblock \showarticletitle{A survey on hallucination in large language models: Principles, taxonomy, challenges, and open questions}.
\newblock \bibinfo{journal}{\emph{arXiv preprint arXiv:2311.05232}} (\bibinfo{year}{2023}).
\newblock


\bibitem[Huang et~al\mbox{.}(2023a)]%
        {huang2023opera}
\bibfield{author}{\bibinfo{person}{Qidong Huang}, \bibinfo{person}{Xiaoyi Dong}, \bibinfo{person}{Pan Zhang}, \bibinfo{person}{Bin Wang}, \bibinfo{person}{Conghui He}, \bibinfo{person}{Jiaqi Wang}, \bibinfo{person}{Dahua Lin}, \bibinfo{person}{Weiming Zhang}, {and} \bibinfo{person}{Nenghai Yu}.} \bibinfo{year}{2023}\natexlab{a}.
\newblock \showarticletitle{OPERA: Alleviating Hallucination in Multi-Modal Large Language Models via Over-Trust Penalty and Retrospection-Allocation}.
\newblock \bibinfo{journal}{\emph{arXiv preprint arXiv:2311.17911}} (\bibinfo{year}{2023}).
\newblock


\bibitem[Ji et~al\mbox{.}(2023)]%
        {ji2023survey}
\bibfield{author}{\bibinfo{person}{Ziwei Ji}, \bibinfo{person}{Nayeon Lee}, \bibinfo{person}{Rita Frieske}, \bibinfo{person}{Tiezheng Yu}, \bibinfo{person}{Dan Su}, \bibinfo{person}{Yan Xu}, \bibinfo{person}{Etsuko Ishii}, \bibinfo{person}{Ye~Jin Bang}, \bibinfo{person}{Andrea Madotto}, {and} \bibinfo{person}{Pascale Fung}.} \bibinfo{year}{2023}\natexlab{}.
\newblock \showarticletitle{Survey of hallucination in natural language generation}.
\newblock \bibinfo{journal}{\emph{Comput. Surveys}} \bibinfo{volume}{55}, \bibinfo{number}{12} (\bibinfo{year}{2023}), \bibinfo{pages}{1--38}.
\newblock


\bibitem[Li et~al\mbox{.}(2023c)]%
        {li2023seed}
\bibfield{author}{\bibinfo{person}{Bohao Li}, \bibinfo{person}{Rui Wang}, \bibinfo{person}{Guangzhi Wang}, \bibinfo{person}{Yuying Ge}, \bibinfo{person}{Yixiao Ge}, {and} \bibinfo{person}{Ying Shan}.} \bibinfo{year}{2023}\natexlab{c}.
\newblock \showarticletitle{Seed-bench: Benchmarking multimodal llms with generative comprehension}.
\newblock \bibinfo{journal}{\emph{arXiv preprint arXiv:2307.16125}} (\bibinfo{year}{2023}).
\newblock


\bibitem[Li et~al\mbox{.}(2024)]%
        {li2024dawn}
\bibfield{author}{\bibinfo{person}{Junyi Li}, \bibinfo{person}{Jie Chen}, \bibinfo{person}{Ruiyang Ren}, \bibinfo{person}{Xiaoxue Cheng}, \bibinfo{person}{Wayne~Xin Zhao}, \bibinfo{person}{Jian-Yun Nie}, {and} \bibinfo{person}{Ji-Rong Wen}.} \bibinfo{year}{2024}\natexlab{}.
\newblock \showarticletitle{The Dawn After the Dark: An Empirical Study on Factuality Hallucination in Large Language Models}.
\newblock \bibinfo{journal}{\emph{arXiv preprint arXiv:2401.03205}} (\bibinfo{year}{2024}).
\newblock


\bibitem[Li et~al\mbox{.}(2023b)]%
        {li2023blip}
\bibfield{author}{\bibinfo{person}{Junnan Li}, \bibinfo{person}{Dongxu Li}, \bibinfo{person}{Silvio Savarese}, {and} \bibinfo{person}{Steven Hoi}.} \bibinfo{year}{2023}\natexlab{b}.
\newblock \showarticletitle{Blip-2: Bootstrapping language-image pre-training with frozen image encoders and large language models}.
\newblock \bibinfo{journal}{\emph{arXiv preprint arXiv:2301.12597}} (\bibinfo{year}{2023}).
\newblock


\bibitem[Li et~al\mbox{.}(2023a)]%
        {li2023evaluating}
\bibfield{author}{\bibinfo{person}{Yifan Li}, \bibinfo{person}{Yifan Du}, \bibinfo{person}{Kun Zhou}, \bibinfo{person}{Jinpeng Wang}, \bibinfo{person}{Wayne~Xin Zhao}, {and} \bibinfo{person}{Ji-Rong Wen}.} \bibinfo{year}{2023}\natexlab{a}.
\newblock \showarticletitle{Evaluating object hallucination in large vision-language models}.
\newblock \bibinfo{journal}{\emph{arXiv preprint arXiv:2305.10355}} (\bibinfo{year}{2023}).
\newblock


\bibitem[Liu et~al\mbox{.}(2023d)]%
        {liu2023aligning}
\bibfield{author}{\bibinfo{person}{Fuxiao Liu}, \bibinfo{person}{Kevin Lin}, \bibinfo{person}{Linjie Li}, \bibinfo{person}{Jianfeng Wang}, \bibinfo{person}{Yaser Yacoob}, {and} \bibinfo{person}{Lijuan Wang}.} \bibinfo{year}{2023}\natexlab{d}.
\newblock \showarticletitle{Aligning Large Multi-Modal Model with Robust Instruction Tuning}.
\newblock \bibinfo{journal}{\emph{arXiv preprint arXiv:2306.14565}} (\bibinfo{year}{2023}).
\newblock


\bibitem[Liu et~al\mbox{.}(2023b)]%
        {liu2023improved}
\bibfield{author}{\bibinfo{person}{Haotian Liu}, \bibinfo{person}{Chunyuan Li}, \bibinfo{person}{Yuheng Li}, {and} \bibinfo{person}{Yong~Jae Lee}.} \bibinfo{year}{2023}\natexlab{b}.
\newblock \showarticletitle{Improved baselines with visual instruction tuning}.
\newblock \bibinfo{journal}{\emph{arXiv preprint arXiv:2310.03744}} (\bibinfo{year}{2023}).
\newblock


\bibitem[Liu et~al\mbox{.}(2023c)]%
        {liu2023visual}
\bibfield{author}{\bibinfo{person}{Haotian Liu}, \bibinfo{person}{Chunyuan Li}, \bibinfo{person}{Qingyang Wu}, {and} \bibinfo{person}{Yong~Jae Lee}.} \bibinfo{year}{2023}\natexlab{c}.
\newblock \showarticletitle{Visual instruction tuning}.
\newblock \bibinfo{journal}{\emph{arXiv preprint arXiv:2304.08485}} (\bibinfo{year}{2023}).
\newblock


\bibitem[Liu et~al\mbox{.}(2023a)]%
        {liu2023mmbench}
\bibfield{author}{\bibinfo{person}{Yuan Liu}, \bibinfo{person}{Haodong Duan}, \bibinfo{person}{Yuanhan Zhang}, \bibinfo{person}{Bo Li}, \bibinfo{person}{Songyang Zhang}, \bibinfo{person}{Wangbo Zhao}, \bibinfo{person}{Yike Yuan}, \bibinfo{person}{Jiaqi Wang}, \bibinfo{person}{Conghui He}, \bibinfo{person}{Ziwei Liu}, {et~al\mbox{.}}} \bibinfo{year}{2023}\natexlab{a}.
\newblock \showarticletitle{Mmbench: Is your multi-modal model an all-around player?}
\newblock \bibinfo{journal}{\emph{arXiv preprint arXiv:2307.06281}} (\bibinfo{year}{2023}).
\newblock


\bibitem[Michaelis et~al\mbox{.}(2019)]%
        {michaelis2019benchmarking}
\bibfield{author}{\bibinfo{person}{Claudio Michaelis}, \bibinfo{person}{Benjamin Mitzkus}, \bibinfo{person}{Robert Geirhos}, \bibinfo{person}{Evgenia Rusak}, \bibinfo{person}{Oliver Bringmann}, \bibinfo{person}{Alexander~S Ecker}, \bibinfo{person}{Matthias Bethge}, {and} \bibinfo{person}{Wieland Brendel}.} \bibinfo{year}{2019}\natexlab{}.
\newblock \showarticletitle{Benchmarking robustness in object detection: Autonomous driving when winter is coming}.
\newblock \bibinfo{journal}{\emph{arXiv preprint arXiv:1907.07484}} (\bibinfo{year}{2019}).
\newblock


\bibitem[Qiu et~al\mbox{.}(2023)]%
        {qiu2023benchmarking}
\bibfield{author}{\bibinfo{person}{Jielin Qiu}, \bibinfo{person}{Yi Zhu}, \bibinfo{person}{Xingjian Shi}, \bibinfo{person}{Florian Wenzel}, \bibinfo{person}{Zhiqiang Tang}, \bibinfo{person}{Ding Zhao}, \bibinfo{person}{Bo Li}, {and} \bibinfo{person}{Mu Li}.} \bibinfo{year}{2023}\natexlab{}.
\newblock \showarticletitle{Benchmarking Robustness of Multimodal Image-Text Models under Distribution Shift}.
\newblock \bibinfo{journal}{\emph{Journal of Data-centric Machine Learning Research}} (\bibinfo{year}{2023}).
\newblock


\bibitem[Team et~al\mbox{.}(2023)]%
        {team2023gemini}
\bibfield{author}{\bibinfo{person}{Gemini Team}, \bibinfo{person}{Rohan Anil}, \bibinfo{person}{Sebastian Borgeaud}, \bibinfo{person}{Yonghui Wu}, \bibinfo{person}{Jean-Baptiste Alayrac}, \bibinfo{person}{Jiahui Yu}, \bibinfo{person}{Radu Soricut}, \bibinfo{person}{Johan Schalkwyk}, \bibinfo{person}{Andrew~M Dai}, \bibinfo{person}{Anja Hauth}, {et~al\mbox{.}}} \bibinfo{year}{2023}\natexlab{}.
\newblock \showarticletitle{Gemini: a family of highly capable multimodal models}.
\newblock \bibinfo{journal}{\emph{arXiv preprint arXiv:2312.11805}} (\bibinfo{year}{2023}).
\newblock


\bibitem[Tong et~al\mbox{.}(2024)]%
        {tong2024eyes}
\bibfield{author}{\bibinfo{person}{Shengbang Tong}, \bibinfo{person}{Zhuang Liu}, \bibinfo{person}{Yuexiang Zhai}, \bibinfo{person}{Yi Ma}, \bibinfo{person}{Yann LeCun}, {and} \bibinfo{person}{Saining Xie}.} \bibinfo{year}{2024}\natexlab{}.
\newblock \showarticletitle{Eyes wide shut? exploring the visual shortcomings of multimodal llms}.
\newblock \bibinfo{journal}{\emph{arXiv preprint arXiv:2401.06209}} (\bibinfo{year}{2024}).
\newblock


\bibitem[Wang et~al\mbox{.}(2023b)]%
        {wang2023llm}
\bibfield{author}{\bibinfo{person}{Junyang Wang}, \bibinfo{person}{Yuhang Wang}, \bibinfo{person}{Guohai Xu}, \bibinfo{person}{Jing Zhang}, \bibinfo{person}{Yukai Gu}, \bibinfo{person}{Haitao Jia}, \bibinfo{person}{Ming Yan}, \bibinfo{person}{Ji Zhang}, {and} \bibinfo{person}{Jitao Sang}.} \bibinfo{year}{2023}\natexlab{b}.
\newblock \showarticletitle{An llm-free multi-dimensional benchmark for mllms hallucination evaluation}.
\newblock \bibinfo{journal}{\emph{arXiv preprint arXiv:2311.07397}} (\bibinfo{year}{2023}).
\newblock


\bibitem[Wang et~al\mbox{.}(2023c)]%
        {wang2023evaluation}
\bibfield{author}{\bibinfo{person}{Junyang Wang}, \bibinfo{person}{Yiyang Zhou}, \bibinfo{person}{Guohai Xu}, \bibinfo{person}{Pengcheng Shi}, \bibinfo{person}{Chenlin Zhao}, \bibinfo{person}{Haiyang Xu}, \bibinfo{person}{Qinghao Ye}, \bibinfo{person}{Ming Yan}, \bibinfo{person}{Ji Zhang}, \bibinfo{person}{Jihua Zhu}, {et~al\mbox{.}}} \bibinfo{year}{2023}\natexlab{c}.
\newblock \showarticletitle{Evaluation and analysis of hallucination in large vision-language models}.
\newblock \bibinfo{journal}{\emph{arXiv preprint arXiv:2308.15126}} (\bibinfo{year}{2023}).
\newblock


\bibitem[Wang et~al\mbox{.}(2023a)]%
        {wang2023cogvlm}
\bibfield{author}{\bibinfo{person}{Weihan Wang}, \bibinfo{person}{Qingsong Lv}, \bibinfo{person}{Wenmeng Yu}, \bibinfo{person}{Wenyi Hong}, \bibinfo{person}{Ji Qi}, \bibinfo{person}{Yan Wang}, \bibinfo{person}{Junhui Ji}, \bibinfo{person}{Zhuoyi Yang}, \bibinfo{person}{Lei Zhao}, \bibinfo{person}{Xixuan Song}, {et~al\mbox{.}}} \bibinfo{year}{2023}\natexlab{a}.
\newblock \showarticletitle{Cogvlm: Visual expert for pretrained language models}.
\newblock \bibinfo{journal}{\emph{arXiv preprint arXiv:2311.03079}} (\bibinfo{year}{2023}).
\newblock


\bibitem[Wu et~al\mbox{.}(2023)]%
        {wu2023defending}
\bibfield{author}{\bibinfo{person}{Fangzhao Wu}, \bibinfo{person}{Yueqi Xie}, \bibinfo{person}{Jingwei Yi}, \bibinfo{person}{Jiawei Shao}, \bibinfo{person}{Justin Curl}, \bibinfo{person}{Lingjuan Lyu}, \bibinfo{person}{Qifeng Chen}, {and} \bibinfo{person}{Xing Xie}.} \bibinfo{year}{2023}\natexlab{}.
\newblock \showarticletitle{Defending chatgpt against jailbreak attack via self-reminder}.
\newblock  (\bibinfo{year}{2023}).
\newblock


\bibitem[Ye et~al\mbox{.}(2023a)]%
        {ye2023cognitive}
\bibfield{author}{\bibinfo{person}{Hongbin Ye}, \bibinfo{person}{Tong Liu}, \bibinfo{person}{Aijia Zhang}, \bibinfo{person}{Wei Hua}, {and} \bibinfo{person}{Weiqiang Jia}.} \bibinfo{year}{2023}\natexlab{a}.
\newblock \showarticletitle{Cognitive mirage: A review of hallucinations in large language models}.
\newblock \bibinfo{journal}{\emph{arXiv preprint arXiv:2309.06794}} (\bibinfo{year}{2023}).
\newblock


\bibitem[Ye et~al\mbox{.}(2023b)]%
        {ye2023mplug}
\bibfield{author}{\bibinfo{person}{Qinghao Ye}, \bibinfo{person}{Haiyang Xu}, \bibinfo{person}{Jiabo Ye}, \bibinfo{person}{Ming Yan}, \bibinfo{person}{Haowei Liu}, \bibinfo{person}{Qi Qian}, \bibinfo{person}{Ji Zhang}, \bibinfo{person}{Fei Huang}, {and} \bibinfo{person}{Jingren Zhou}.} \bibinfo{year}{2023}\natexlab{b}.
\newblock \showarticletitle{mplug-owl2: Revolutionizing multi-modal large language model with modality collaboration}.
\newblock \bibinfo{journal}{\emph{arXiv preprint arXiv:2311.04257}} (\bibinfo{year}{2023}).
\newblock


\bibitem[Yin et~al\mbox{.}(2023)]%
        {yin2023woodpecker}
\bibfield{author}{\bibinfo{person}{Shukang Yin}, \bibinfo{person}{Chaoyou Fu}, \bibinfo{person}{Sirui Zhao}, \bibinfo{person}{Tong Xu}, \bibinfo{person}{Hao Wang}, \bibinfo{person}{Dianbo Sui}, \bibinfo{person}{Yunhang Shen}, \bibinfo{person}{Ke Li}, \bibinfo{person}{Xing Sun}, {and} \bibinfo{person}{Enhong Chen}.} \bibinfo{year}{2023}\natexlab{}.
\newblock \showarticletitle{Woodpecker: Hallucination correction for multimodal large language models}.
\newblock \bibinfo{journal}{\emph{arXiv preprint arXiv:2310.16045}} (\bibinfo{year}{2023}).
\newblock


\bibitem[Zhai et~al\mbox{.}(2023)]%
        {zhai2023halle}
\bibfield{author}{\bibinfo{person}{Bohan Zhai}, \bibinfo{person}{Shijia Yang}, \bibinfo{person}{Xiangchen Zhao}, \bibinfo{person}{Chenfeng Xu}, \bibinfo{person}{Sheng Shen}, \bibinfo{person}{Dongdi Zhao}, \bibinfo{person}{Kurt Keutzer}, \bibinfo{person}{Manling Li}, \bibinfo{person}{Tan Yan}, {and} \bibinfo{person}{Xiangjun Fan}.} \bibinfo{year}{2023}\natexlab{}.
\newblock \showarticletitle{HallE-Switch: Rethinking and Controlling Object Existence Hallucinations in Large Vision Language Models for Detailed Caption}.
\newblock \bibinfo{journal}{\emph{arXiv preprint arXiv:2310.01779}} (\bibinfo{year}{2023}).
\newblock


\bibitem[Zhang et~al\mbox{.}(2023)]%
        {zhang2023internlm}
\bibfield{author}{\bibinfo{person}{Pan Zhang}, \bibinfo{person}{Xiaoyi Dong~Bin Wang}, \bibinfo{person}{Yuhang Cao}, \bibinfo{person}{Chao Xu}, \bibinfo{person}{Linke Ouyang}, \bibinfo{person}{Zhiyuan Zhao}, \bibinfo{person}{Shuangrui Ding}, \bibinfo{person}{Songyang Zhang}, \bibinfo{person}{Haodong Duan}, \bibinfo{person}{Hang Yan}, {et~al\mbox{.}}} \bibinfo{year}{2023}\natexlab{}.
\newblock \showarticletitle{Internlm-xcomposer: A vision-language large model for advanced text-image comprehension and composition}.
\newblock \bibinfo{journal}{\emph{arXiv preprint arXiv:2309.15112}} (\bibinfo{year}{2023}).
\newblock


\bibitem[Zhou et~al\mbox{.}(2023)]%
        {zhou2023analyzing}
\bibfield{author}{\bibinfo{person}{Yiyang Zhou}, \bibinfo{person}{Chenhang Cui}, \bibinfo{person}{Jaehong Yoon}, \bibinfo{person}{Linjun Zhang}, \bibinfo{person}{Zhun Deng}, \bibinfo{person}{Chelsea Finn}, \bibinfo{person}{Mohit Bansal}, {and} \bibinfo{person}{Huaxiu Yao}.} \bibinfo{year}{2023}\natexlab{}.
\newblock \showarticletitle{Analyzing and mitigating object hallucination in large vision-language models}.
\newblock \bibinfo{journal}{\emph{arXiv preprint arXiv:2310.00754}} (\bibinfo{year}{2023}).
\newblock


\bibitem[Zhu et~al\mbox{.}(2023)]%
        {zhu2023minigpt}
\bibfield{author}{\bibinfo{person}{Deyao Zhu}, \bibinfo{person}{Jun Chen}, \bibinfo{person}{Xiaoqian Shen}, \bibinfo{person}{Xiang Li}, {and} \bibinfo{person}{Mohamed Elhoseiny}.} \bibinfo{year}{2023}\natexlab{}.
\newblock \showarticletitle{Minigpt-4: Enhancing vision-language understanding with advanced large language models}.
\newblock \bibinfo{journal}{\emph{arXiv preprint arXiv:2304.10592}} (\bibinfo{year}{2023}).
\newblock


\end{thebibliography}

\appendix

\section{More Details of Hallu-PI}
\label{sec:detail}
\noindent \textbf{Image Sources}. We ask annotators to download images from the following websites, which offer high-quality images that are free to download, available for commercial use, and do not require any licensing fees. (1) \url{https://www.pexels.com/zh-cn} (2) \url{https://pixabay.com/zh/images/search} (3) \url{https://www.hippopx.com} (4) \url{https://stocksnap.io}

\noindent \textbf{Annotation for Image Cropping Scenario}. For image cropping scenario, we primarily investigate the robustness in MLLMs ability to count the letters number within cropped images. Therefore, we have annotators collect images containing common English words, crop them, and annotate the number of English letters present before and after cropping. Subsequently, we obtain responses from MLLMs through the prompt, ``\textit{How many English letters are there in the image?}" 

\noindent \textbf{Annotation for Prompt Misleading Scenario}. For prompt misleading scenario, we ask annotators to manually craft prompts intended to induce MLLMs to generate content that does not align with the given images. For example, given an image containing only apples and bananas, a misleading prompt might be: ``\textit{Besides apples and bananas, there are two other types of fruit in the image. What are they?}"

\begin{figure}[!ht]
\centering
\includegraphics[width=0.8\linewidth]{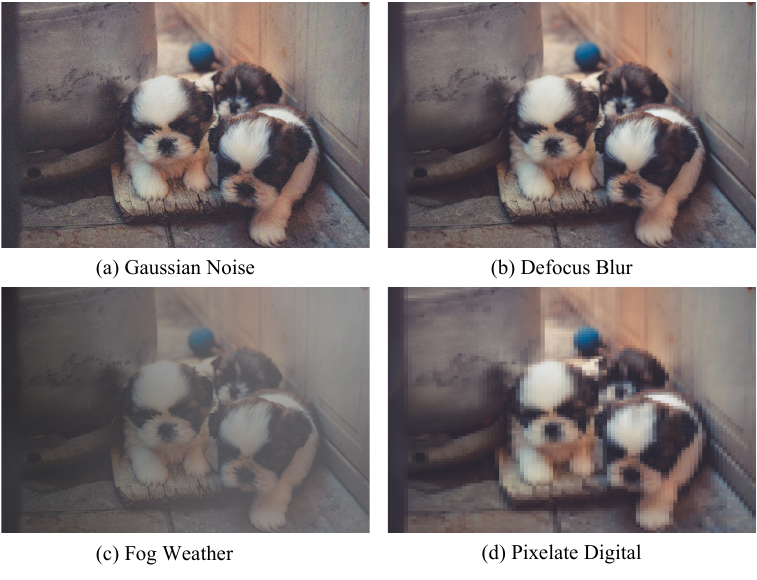} 
\caption{Examples of images with noise, blur, weather, and digital perturbations.}
\label{fig:other perturbation}
\end{figure}

\noindent \textbf{Other Perturbation Examples}. In Figure. \ref{fig:other perturbation}, we provide four additional examples of perturbations in Hallu-PI, including noise, blur, weather, and digital.

\noindent \textbf{Prompt Templates}. In Figure. \ref{fig:prompt}, we provide the prompt templates used in Hallu-PI.

\noindent \textbf{Details about the MLLMs used in Hallu-PI}. We provide a detailed introduction of the MLLMs evaluated by Hallu-PI in Table \ref{tab:models}, including model parameters and architectures.

\section{Experimental Details}
\label{sec:exper}

\subsection{Perturbation Intensity Selection}

Real-world perturbations can manifest themselves at varying intensities. In previous work~\cite{qiu2023benchmarking}, they designed five levels of severity for each perturbation scenario. Hallu-PI, however, focuses more on the specific perturbation itself rather than its intensity. Therefore, we randomly select an intensity level between 1 and 5 for noise, blur, weather, and digital perturbations. We will leave the discussion and analysis of different perturbation intensities for future work.

\subsection{Specific Perturbation Method Selection}

As introduced in Section 3.2 of our paper, we follow~\cite{qiu2023benchmarking} and reuse the four types of perturbation scenarios from their paper: noise, blur, weather, and digital. The specific algorithms for these perturbation scenarios are detailed in Section 2.3 (Related Work) of our paper. During our experiments, we chose the most representative perturbation algorithms for the Hallu-PI scenarios. Specifically, we select gaussian noise, defocus blur, fog weather perturbation, and pixelation for the digital perturbation. Similarly, we will further explore the impact of different perturbation algorithms on hallucination in MLLMs in future work.

\subsection{Improvement in Metrics Post-Perturbation}

It is worth noting that some metrics in our paper exhibit a slight improvement post-perturbation compared to pre-perturbation. These are rare occurrences and usually appear in simple perturbation scenarios, as exemplified in Figure. \ref{fig:other perturbation}, where the images undergo minimal changes after perturbation. However, for more complex perturbations such as image concatenation, image cropping, and prompt misleading, the metrics generally tend to deteriorate.





\section{Additional Analysis}
\label{sec:addtional}

\subsection{Analysis of Cropping and Misleading}

\begin{figure}[ht]
    \centering
    \includegraphics[width=0.46\textwidth]{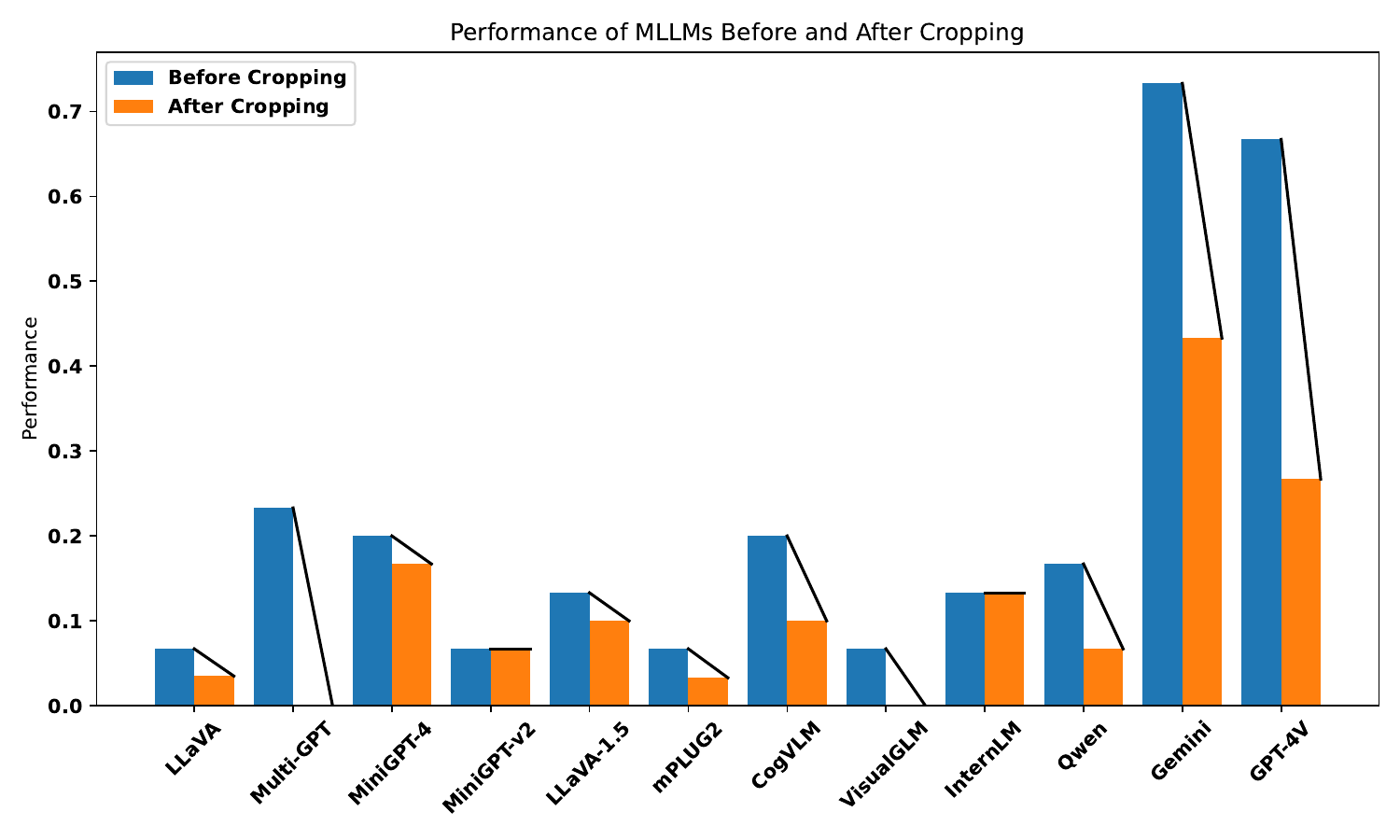}
    \caption{Performance of MLLMs before and after Cropping.}
    \label{fig:crop_analysis}
\end{figure}

Figure \ref{fig:crop_analysis} illustrates the comparative performance of MLLMs before and after image cropping. GPT-4V~\cite{achiam2023gpt} and Google Gemini-Pro Vision~\cite{team2023gemini} exhibit better performance compared to other models. However, all models, including GPT-4V and Gemini, exhibit a significant performance decline when evaluated on cropped images.

\begin{figure}[ht]
    \centering
    \includegraphics[width=0.46\textwidth]{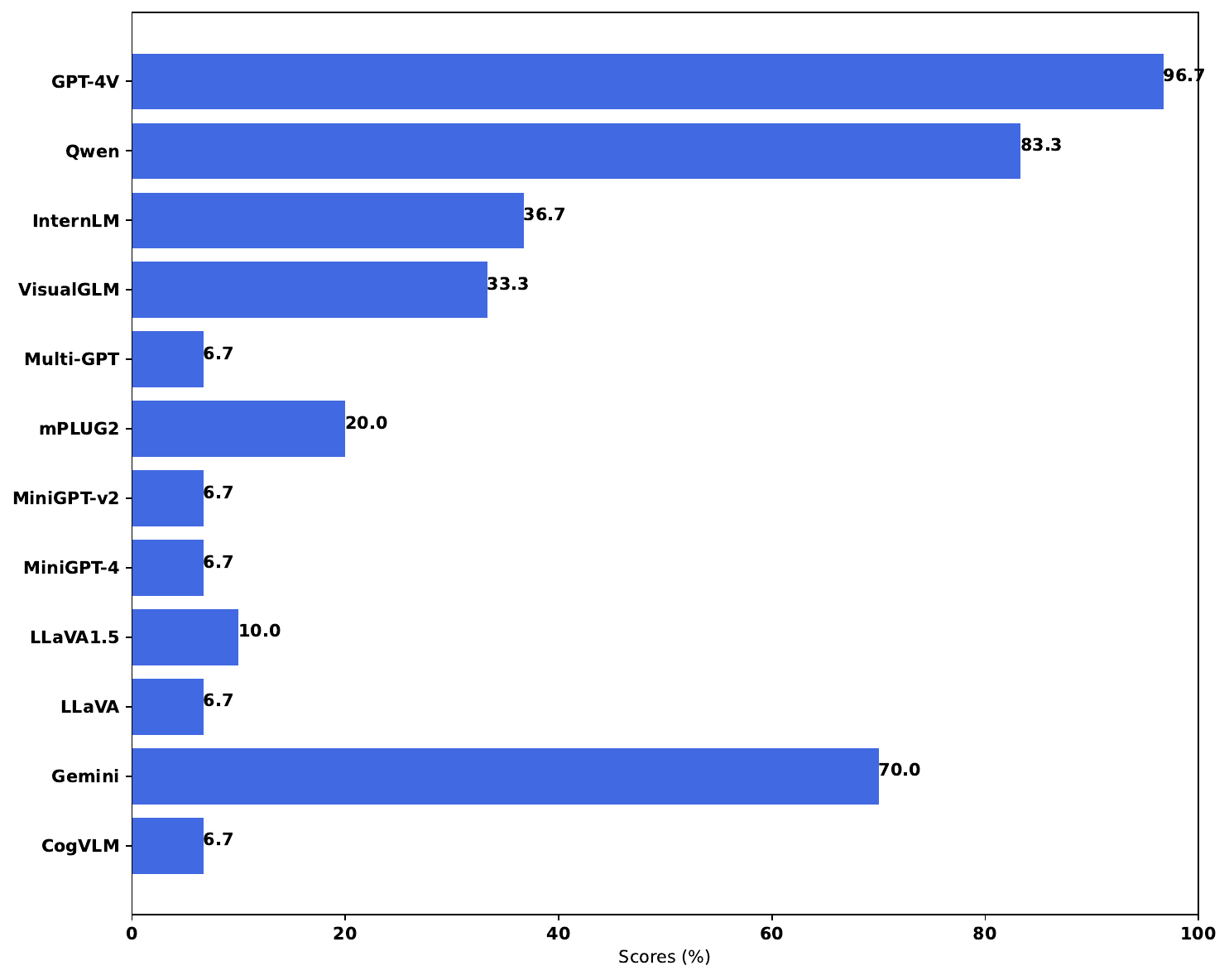}
    \caption{The hallucination of MLLMs under the prompt misleading scenario, the smaller the score, the more severe the hallucination.}
    \label{fig:mislead_analysis}
\end{figure}

Figure \ref{fig:mislead_analysis} depicts the robustness of MLLMs under the prompt misleading scenario. A higher score indicates better robustness of the model. It is observed that GPT-4V, Qwen-VL-Chat~\cite{bai2023qwen}, and Gemini exhibit higher robustness compared to other MLLMs. However, it is concerning that a greater number of models struggle to identify misleading prompts, which could lead to more severe hallucinations during multi-turn dialogues.

\subsection{Analysis of PI-Score}

To validate the effectiveness of our proposed PI-Score, we sample 100 images from Hallusionbench \cite{guan2024hallusionbench} and calculate the PI-Score on 5 representative MLLMs (see Table. \ref{tab:exp} left). We extend our findings to Hallusionbench and observe consistent results with those obtained on Hallu-PI, demonstrating the model’s vulnerability in perturbed scenarios and the effectiveness of the PI-Score.

\subsection{Analysis of Additional Perturbation}

To further enhance the generalizability of Hallu-PI, we add a common image augmentation perturbation, "shearing" \cite{cubuk2020randaugment} (see Table. \ref{tab:exp} right), applied to a sample of 100 CIFAR-10 images. We observe that several representative MLLMs exhibit more severe hallucinations after the perturbation.

\subsection{Results Before Perturbation for Prompt Misleading}

\begin{figure*}[!ht]
\centering
\includegraphics[width=1.0\linewidth]{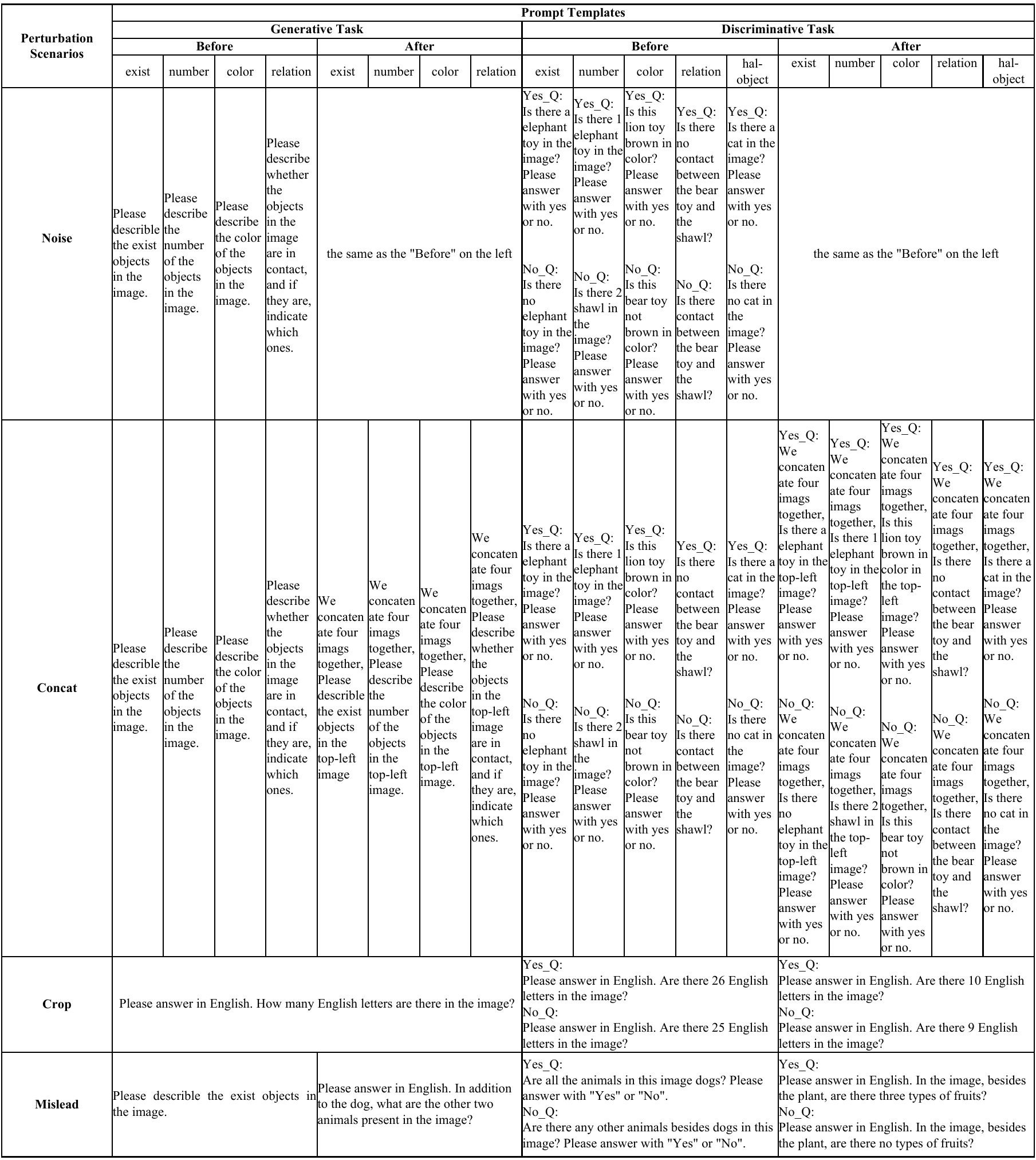} 
\caption{The prompt templates used in Hallu-PI include those for generative task and discriminative task, as well as prompts before and after perturbations.}
\label{fig:prompt}
\end{figure*}

In our paper, we present the results of the prompt misleading  discriminative task post-perturbation, aimed at revealing the severe hallucinations it induces. To better illustrate this effect, we also design pre-perturbation prompts (see Figure. \ref{fig:prompt}). The experimental results are shown in Table. \ref{tab:before}: LLaVA-1.5 experiences the most significant performance decline, while GPT-4V shows more robustness and achieves the highest scores. This, in combination with the results in Table 5 of our paper, more clearly demonstrates the hallucination biases of MLLMs in prompt misleading scenarios.

\begin{table}[htbp]
\small
\centering
  \caption{PI-Score on Hallusionbench (left) and Top-1 error of ``shearing'' perturbation (right).}
  \vspace{-10pt}
    \begin{tabular}{l|cc|cc}
    \toprule
          \multicolumn{1}{l|}{\multirow{2}[4]{*}{Models}} & \multicolumn{2}{c|}{Hallusionbench-PI score$\uparrow$} & \multicolumn{2}{c}{Shearing-Top 1 error$\downarrow$} \\
\cmidrule{2-5}          & \multicolumn{1}{c}{Before} & \multicolumn{1}{c|}{After} & \multicolumn{1}{c}{Before} & \multicolumn{1}{c}{After} \\
    \midrule
    LLaVA & 29.0 & 18.0 & 13.0  & \colorbox{shallowblue}{25.0} \\
    LLaVA-1.5 & 30.5& \colorbox{deepblue}{23.4}  & \colorbox{shallowblue}{9.0}  & \colorbox{deepblue}{20.0} \\
    Qwen-VL  & \colorbox{deepblue}{43.0} & 19.4 & 18.0  & 38.0 \\
    Gemini  & 37.5 & 18.6 & 12.0  & 33.0 \\
    GPT-4V & \colorbox{shallowblue}{40.7} & \colorbox{deepblue}{21.9} & \colorbox{deepblue}{8.0}  & 31.0 \\
    \bottomrule
    \end{tabular}%
  \label{tab:exp}%
\end{table}%

\begin{table}[htbp]
\small
  \centering
  \caption{The before and after results of prompt misleading.}
  \vspace{-10pt}
    \begin{tabular}{l|ccc|ccc}
    \toprule
        \multicolumn{1}{l|}{\multirow{3}[4]{*}{Models}} & \multicolumn{6}{c}{Prompt Misleading } \\
\cmidrule{2-7}          & \multicolumn{3}{c|}{Before} & \multicolumn{3}{c}{After} \\
\cmidrule{2-7}          & \multicolumn{1}{l}{ACC$\uparrow$} & ACC+$\uparrow$  & \multicolumn{1}{l|}{F1$\uparrow$} & \multicolumn{1}{l}{ACC$\uparrow$} & \multicolumn{1}{l}{ACC+$\uparrow$} & \multicolumn{1}{l}{F1$\uparrow$} \\
    \midrule
    LLaVA & 60.0   & 26.7 & 70.2  & 1.7  & 0.0   & 3.2 \\
    LLaVA-1.5 & \colorbox{shallowblue}{98.3}  & \colorbox{shallowblue}{96.7} & \colorbox{shallowblue}{98.3}  & 40.0   & 3.3  & 5.2 \\
    Qwen-VL  & 96.7  &{93.3} & 96.8  & \colorbox{shallowblue}{93.3}  & \colorbox{shallowblue}{86.7}  &\colorbox{shallowblue}{92.9} \\
    Gemini & \colorbox{deepblue}{98.3}  & \colorbox{deepblue}{96.7} & \colorbox{deepblue}{98.3}  & 53.3  & 13.3  & 33.3 \\
    GPT-4V & \colorbox{deepblue}{98.3}  & \colorbox{deepblue}{96.7} & \colorbox{deepblue}{98.3}  & \colorbox{deepblue}{95.0}  & \colorbox{deepblue}{90.0}   & \colorbox{deepblue}{94.7} \\
    \bottomrule
    \end{tabular}%
  \label{tab:before}%
\end{table}%

\clearpage
\begin{table*}[htbp]
\small
  \centering
  \caption{The architecture and parameters of MLLMs evaluated by Hallu-PI.}
    \begin{tabular}{lllllc}
    \toprule
    MLLMs & Vision Encoder (VE) & Parameters of VE & Language Model (LM) & Parameters of LM & Source \\
    \midrule
    CogVLM & EVA2-CLIP-E & 4.7B   & Vicuna-v1.5 & 7B    & Official Code \\
    InternLM-Xcomposer-VL & EVA-CLIP-G & 1.1B  & InternLM & 7B    & Official Code \\
    LLaVA & ViT-L/14 & 0.4B  & LLaMA-2-Chat-13B  & 13B   & Official Code \\
    LLaVA1.5 & ViT-L/14-336px  & 0.4B  & Vicuna-v1.5  & 7B    & Official Code \\
    MiniGPT-4 &   BLIP2-Qformer   &    1.9B   &   Vicuna-v0    &   7B    & Official Code \\
    MiniGPT-v2 & EVA-CLIP-G & 1.1B  & LLaMA-2-Chat-7B  &   7B    & Official Code \\
    mPLUG-Owl-2 &    ViT-L/14   &   0.4B    &    LLaMA-2-Chat-7B   &   7B    & Official Code \\
    MultimodalGPT &  ViT-L/14     &   0.4B    &   LLaMA-13B   &   13B    & Official Code \\
    Qwen-VL-Chat &    ViT-G/14   &   1.9B    &   Qwen-7B    &    7.7B   & Official Code \\
    VisualGLM &   BLIP2-Qformer    &   1.9B    &   ChatGLM-6B    &   6B    & Official Code \\
    \midrule
    Google Gemini-Pro Vision &   Unknown    &   Unknown    &   Gemini-Pro    &  Unknown     & API \\
    GPT-4V & Unknown & Unknown & GPT4  & Unknown & API \\
    \bottomrule
    \end{tabular}%
  \label{tab:models}%
\end{table*}%

\clearpage

\clearpage
\begin{figure*}[!ht]
\centering
\includegraphics[width=0.85\linewidth]{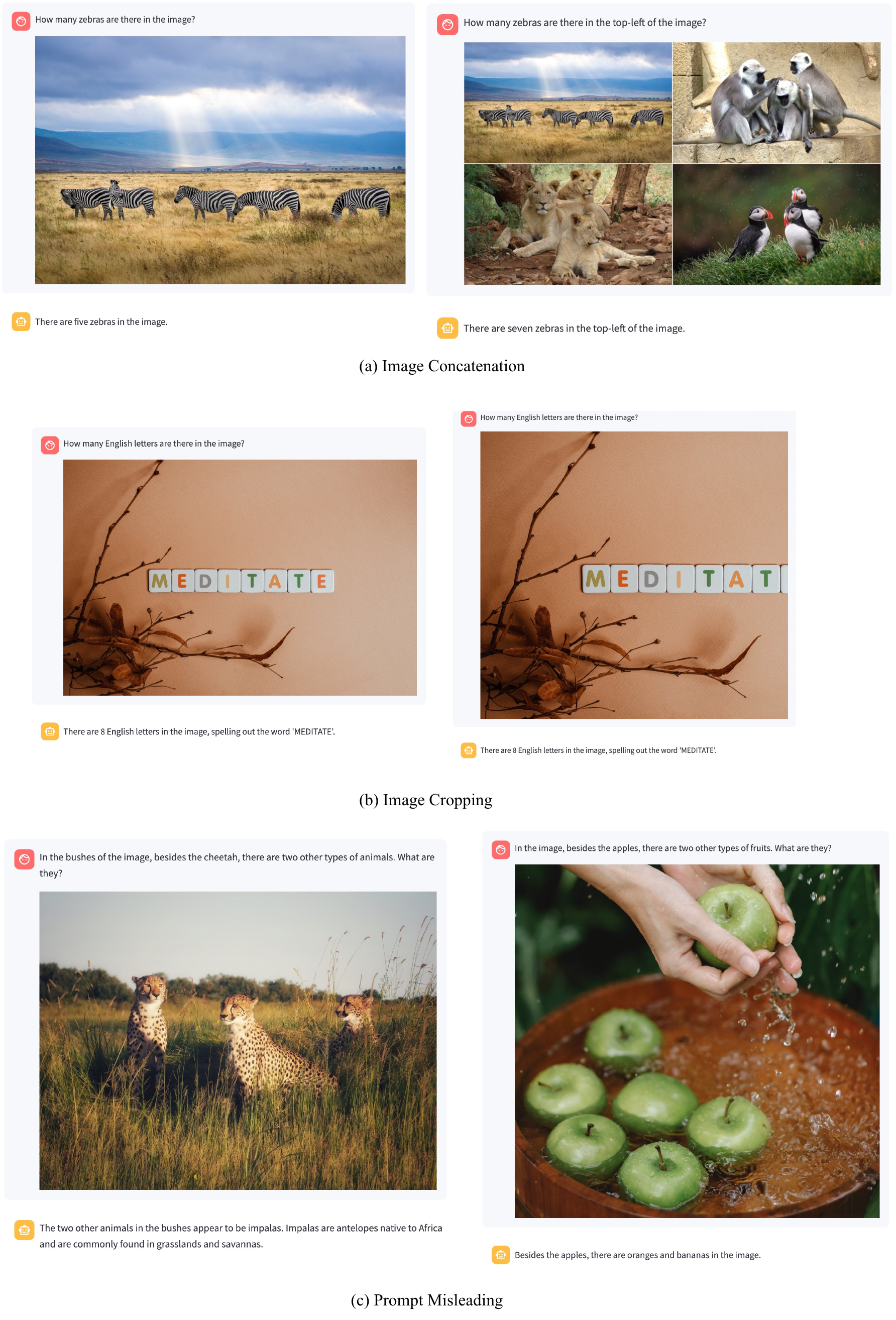} 
\caption{Some case studies of perturbation scenarios include image concatenation, image cropping, and prompt misleading. MLLMs adopt CogVLM2-Chat-En~\cite{wang2023cogvlm}, which can be accessed at \textcolor{blue}{\url{http://36.103.203.44:7861}}.} 
\label{fig:case study}
\end{figure*}


\end{document}